\documentclass{article}

\usepackage[nonatbib,final]{neurips_2020}

\usepackage[utf8]{inputenc} %
\usepackage[T1]{fontenc}    %
\usepackage{hyperref}       %
\usepackage{url}            %
\usepackage{booktabs}       %
\usepackage{amsfonts}       %
\usepackage{nicefrac}       %
\usepackage{microtype}      %

\usepackage{algorithm, algpseudocode}
\usepackage{graphicx}
\usepackage{subfig}
\usepackage{wrapfig}
\newcommand\blfootnote[1]{%
  \begingroup
  \renewcommand\thefootnote{}\footnote{#1}%
  \addtocounter{footnote}{-1}%
  \endgroup
}

\title{Refactoring Policy for Compositional Generalizability using Self-Supervised Object Proposals}

\author{%
    Tongzhou Mu$^{1*}$
    \quad Jiayuan Gu$^{1*}$
    \quad Zhiwei Jia$^{1}$
    \quad Hao Tang$^{2}$
    \quad Hao Su$^{1}$ \\
    \hspace{2.2em}
    $^{1}$University of California, San Diego
    \hspace{1.3em}
    $^{2}$Shanghai Jiao Tong University\\
    \texttt{\{t3mu,jigu,zjia,haosu\}@eng.ucsd.edu}
    \hspace{0.8em}
    \texttt{tanghaosjtu@gmail.com}
}

\usepackage{subfig}
\usepackage{xcolor}

\definecolor{darkred}{rgb}{0.55, 0.0, 0.0}

\definecolor{dimgray}{rgb}{0.41, 0.41, 0.41}

\newcommand{\eg}{\textit{e}.\textit{g}.}
\begin{document}

\maketitle

\begin{abstract}
We study how to learn a policy with compositional generalizability. We propose a two-stage framework, which \emph{refactorizes} a high-reward teacher policy into a generalizable student policy with strong inductive bias. Particularly, we implement an object-centric GNN-based student policy, whose input objects are learned from images through self-supervised learning.
Empirically, we evaluate our approach on four difficult  tasks that require compositional generalizability, and achieve superior performance compared to baselines.

\end{abstract}

\blfootnote{* These authors contributed equally to this work.}
\blfootnote{Project website: \url{https://jiayuan-gu.github.io/policy-refactorization}.}

\section{Introduction}
Obtaining policies that would work in different environments is a fundamental challenge of artificial intelligence.
While reinforcement learning and imitation learning have made remarkable progress to solve challenging tasks~(e.g., Go~\cite{silver2016mastering}, Montezuma's Revenge~\cite{ecoffet2019go}), their cross-environment generalizability is still limited, affecting applications in complicated and changing scenarios. In fact, to enable such applications, the policy function has to be adaptive to many factors, such as nuisances in visual appearance and physical attributes, change of rewards, and variation in the quantity and arrangement of objects. 

In this paper, we aim to learn a generalizable policy network across environments composed by a flexible number of objects. To make the problem tractable, we further assume that policies of \textbf{consistency} across environments can be discovered, in the sense that a policy with consistent behavior achieves high cumulative reward following the same reasoning rationale in different environments. While this statement lacks rigor mathematically, it describes a common use case in our daily experiences: Take the classical Pacman game for example -- it is not hard to write down a greedy policy that achieves high rewards, no matter how many dots and ghosts there are and where they are located.

We notice that, when talking about policy learning, we often aim to achieve two goals that are both quire challenging but not exactly aligned: 1) to address the \textbf{optimization challenge} for maximizing the reward in the training environments, which often involves high variance due to the randomness of environment dynamics and exploration strategy; and 2) to address the \textbf{generalization challenge}, which aims at achieving desired cross-environment generalizability. We conjecture that solving the two misaligned goals independently may result in two easier problems than solving the coupled problem, and there are richer options to solve each problem independently. 

For this reason, we explore a strategy to break the policy learning into two stages in the compositional generalizability setup: The first stage focuses on overfitting the training environments with no generalizability concern, while the second stage focuses on designing the policy network architecture to compositionally generalize the solution from the first stage. In practice, we can use the teacher-student network distillation idea to train a student network that imitates the teacher policy but has stronger generalizability. To emphasize the goal of enhancing generalizability instead of reducing network size, we call this two-stage learning strategy as \textbf{policy refactorization}, in analogy to code refactorization that aims to boost the reusability of codes. This design relieves the requirement of the first stage. We can choose networks with strong overfitting ability and fast running speed for policy optimization. For the second stage, we design or search network architectures with strong inductive-bias for generalizability. 

Since the generalizability is only expected for the student network, we put more effort in investigating the performance of different student network choices (e.g., CNN, RelationNet~\cite{zambaldi2018deep}, GNN). 
Particularly, we study GNNs with object-centric scene graphs as inputs, driven by the intuition that policy acting upon our physical world should be based on objects, attributes, and the relationship between objects, and that GNNs have strong algorithmic alignment with dynamic programming. However, unlike CNNs and RelationNet, GNNs rely on object bounding boxes to build the underlying scene graph. %
To make the comparison more fair and better connected with the state-of-the-art in object bounding box discovery, we evaluate GNN student policies using self-supervised object detectors learned by our improved SPACE model~\cite{lin2020space}.

We empirically show that overfitting RL policies or heuristic policies can be refactorized into policies with strong compositional generalizability, in a few environments that include flexible number of objects, random object arrangement, and composition of foreground/background. Particularly, in difficult environments with sophisticated reasoning, long-range interaction, or unfamiliar background, the GNN student policy using self-supervised object detector shows the most promising results.

\section{Approach}
\label{sec:approach}
\subsection{Overview}

In this section, we describe our two-stage framework to learn a generalizable policy. In the first stage (Sec~\ref{sec:demonstration-acquisition}), our focus is to address the optimization challenge, that is, we strive to acquire a consistent teacher policy that only needs to perform well in the training environments. And this teacher policy is used to generate demonstration dataset. In the second stage (Sec~\ref{sec:refactorization}), we focus on addressing the generalization challenge. We describe how we refactor the teacher policy into a student policy with strong inductive bias for generalization. Particularly, we study how to implement a GNN-based student policy that is based on a self-supervised high-recall object detector. As a by-product, attributes of objects naturally emerge in the object feature space (Sec~\ref{sec:eoc}).

\subsection{Stage I: Demonstration Acquisition without Generalizability Concerns}
\label{sec:demonstration-acquisition}
In this stage, we have access to the training environment that is interactive and has image-stream based states. The outcome is a demonstration dataset, which contains state-action pairs from a policy achieving high reward in the training environment. To be more specific, we aim to generate a demonstration dataset $\mathcal{D}=\{(I_i, \pi_i)\}_{i=1}^N$, where $I_i$ is the input image (or a stack of several images) from the training environments, and $\pi_i$ is the output of demonstration policy on the input $I_i$. The representation of $\pi_i$ is flexible: it can be an action, the logits, or any latent representation, which indicates the demonstration action distribution. This dataset will be used for training both object detector and GNN-based policy.

We call the policy used to generate demonstration as \textit{teacher policy}. It can be obtained in any way, as long as it provides reasonable and good supervision on how to solve the task, and it is not necessary to have compositional generaliziabilty. For example, one can use reinforcement learning to learn a policy, or can also use a heuristic algorithm to search for a policy. There are three caveats here. First, in our compositional generalizability setup, we usually assume that the training environments have a small number of objects, but the test environments may have many objects. For expensive optimization techniques whose complexity grow fast w.r.t. the number of objects, the setup still allows us to use them in training environments. We expect that the learned policy can scale up to the test environments. Second, besides we expect the policy to have high reward, we are also concerned about the consistency of the policy behavior across environments, and the learnability of the policy function. We do not have a deep understanding of this consistency yet; however, some examples may be helpful to see the point. For example, in the Pacman game with no ghosts, a greedy policy that always approaches the closest dot has consistent behavior across environments, and this greedy policy is learnable by a GNN with edge convolution. More examples can be found in the experiment section. Finally, since we focus on maximizing reward rather than generalizability in the training environment, networks of high-capacity and fast running speed are favorable.

In practice, if we learn the teacher policy by reinforcement learning, we can try (or search) different network architectures (\eg, CNN, GNN) and RL algorithms (\eg, REINFORCE~\cite{sutton_book}, DQN~\cite{mnih2015human}, PPO~\cite{schulman2017proximal}), until we obtain a model solving the training environment best.
The best teacher policy is used to generate the demonstration dataset by interacting with the environment.

\subsection{Stage II: Generalizable Policy Refactorization}
\label{sec:refactorization}

This stage focuses on finding or designing the student network that has good generalizability. In this work, we consider three alternatives: plain CNN, RelationNet, and GNN.
Particularly, we conjecture that GNNs with object-centric scene graphs provide the strongest inductive bias for compositional generalization.
To this end, we implement an effective approach to learn an object detector in a self-supervised manner (Sec~\ref{sec:unsupervised_detection}) and use GNNs to learn policy based on imperfect proposals generated by the object detector (Sec~\ref{sec:gnn_learning}).

\subsubsection{Obtaining High-Recall Object Proposals}
\label{sec:unsupervised_detection}
In this paper, we build a self-supervised object detector based upon SPACE~\cite{lin2020space}.
Different from other self-supervised object detection methods based on reconstruction~\cite{eslami2016attend,crawford2019spatially} or motion~\cite{kosiorek2018sequential,goel2018unsupervised}, SPACE is able to detect salient objects from relative complex backgrounds.
It is self-supervised by reconstructing a single image with foreground objects and background stuff.
However, it is sometimes unstable to train and sensitive to several hyperparameters, \eg, prior object sizes~\cite{github2020space}.
Therefore, we improve SPACE by introducing a better parameterization for bounding box regression, which is widely used in supervised object detection~\cite{girshick2014rich}.
More details can be found in the supplementary.

The reason to use an unsupervised object detector is that it requires no extra labeling, and can be easily applied to widely used environments, \eg, Atari~\cite{bellemare2013arcade} and ProcGen~\cite{cobbe2019procgen}.
However, note that our framework is not designed for any specific object detector.
The only expectation is that the recall should be relatively high with a reasonable precision, which is a practical assumption.

\subsubsection{Learning GNN-based Policy with Objects from Demonstration}
\label{sec:gnn_learning}
The desired compositional generalizability of policy in this work is indeed the consequence of having an underlying reasoning mechanism (i.e., the program to generate a feasible policy) that is invariant to the composition of objects in environments. For example, in a Pacman game, policies with reasonable performance may be derived based on the reasoning of approaching a close dot (food) and escaping nearby ghosts, applicable to different amount of food and ghosts. To achieve such compositional generalizability, we need a tool that has the representation power of capturing such underlying reasoning mechanism (\eg, a greedy algorithm).
Under the assumption that the task is relevant to objects and the relationship between them, we employ graph neural networks to represent the policy based on the detected objects, due to the impressive power of GNN in reasoning, including approximating dynamic programs in certain scenarios~\cite{xu2019can}. 

\paragraph{Behavior Cloning by GNN} Given the demonstration dataset $\mathcal{D}$, we want to learn a GNN-based policy via behaviour cloning \cite{ross2011reduction}. 
Concretely, we minimize the following loss:
$L = \sum_{(I_i, \pi_i)\in{\mathcal{D}}} \Vert f_{\textrm{GNN}}(\mathcal{O}_i, I_i)-\pi_i \Vert^2_2$
, where for each image $I_i$ in the training set, $\mathcal{O}_i$ corresponds to the bounding boxes generated by the object detector, and $f_{\textrm{GNN}}$ denotes the policy GNN to be trained.
The input to our GNN-based policy is an object-centric graph. The nodes are detected objects and the node features are the image patches cropped from the original image based on the object bounding boxes provided by the object detector. And we further encode them into latent features by a CNN. 
As for the edges between the nodes, we can adopt different types of graph structures depending on the type of the task, e.g., complete graph or empty graph (no edges).
The design of our policy GNN architecture will be elaborated in Sec~\ref{sec:exp} in a task-specific manner.

\paragraph{Down-weighting Demonstrations with Incomplete Task-Relevant Object Detection}
Due to the imperfectness of object detector, chances are that some task-relevant objects are missing and task-irrelevant objects are included.
We observe that GNN is robust to irrelevant objects, but is sensitive to incomplete proposals.
It is sometimes impossible to solve certain tasks when any object is missing.
Those bad data samples with incomplete proposals may prevent the policy GNN to learn a good policy as they force the network to sacrifice the generalizability to fit out-of-distribution data.

To this end, we introduce data parameters~\cite{saxena2019data} to downweight data samples with incomplete object proposals.
Different from \cite{saxena2019data}, we apply data parameters to a general loss rather than cross-entropy loss.
Concretely, we associate a data parameter $\sigma_i$ with each data sample $x_i$.
Within a batch, we reweight the total loss according to data parameters:
$L = \sum_{i=1}^{b} \frac{e^{\sigma_i}}{\sum_{j=1}^{b} e^{\sigma_j}} l(x_i)$
, where $l(x_i)$ is the loss calculated on the data sample $x_i$.
It can be also considered as a deep version of RANSAC~\cite{fischler1981random}.

\subsection{Task-Relevant Knowledge Discovery of Objects and Attributes}
\label{sec:eoc}
One advantage of our framework is that it supports interpretable model diagnosis, thanks to the modular design and explicit object modeling.
The task-relevant feature is learned for each object by our policy GNN.
We can apply K-means clustering or other data mining techniques to learned object features to discover task-relevant knowledge.
Object attributes emerge through the clustering process, and other post-processing can be applied to remove task-irrelevant objects or label data by clustering results. We show some qualitative results in Sec \ref{sec:exp}. 
\section{Experiments}
\label{sec:exp}
We experiment on four different environments to evaluate our proposed methods. We start from evaluating the basic units, the SPACE object detector and the object-centric GNN on Multi-MNIST. After the units are verified, then, we evaluate the effectiveness of our framework for two types of compositional generalizability: w.r.t. the change of object quantity (FallingDigit), and w.r.t. the change of background (BigFish). Finally, we show that there exist environments, \eg, Pacman, in which a generalizable student policy does not have to be object-centric GNNs. 

\subsection{Multi-MNIST}
\label{sec:exp-multi-mnist}

\begin{figure}[t]
\centering
\subfloat[Multi-MNIST-CIFAR]{
\includegraphics[width=0.12\textwidth]{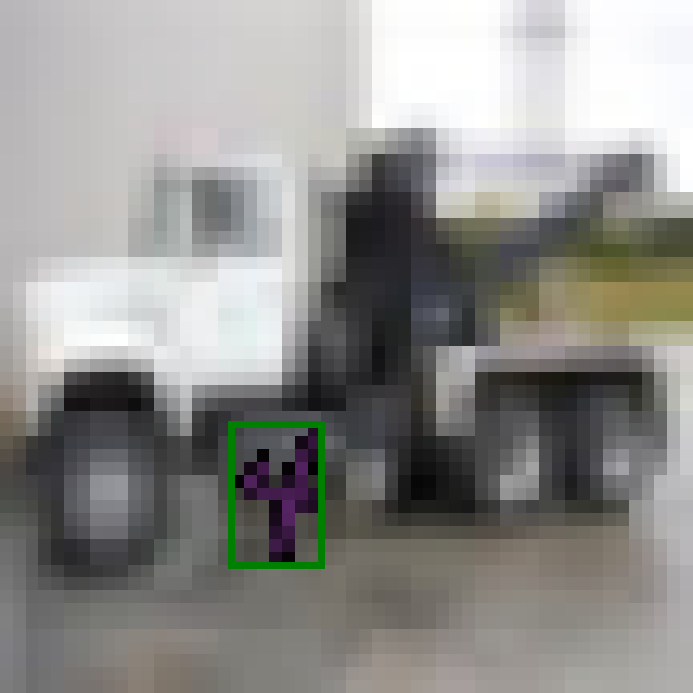}
\includegraphics[width=0.12\textwidth]{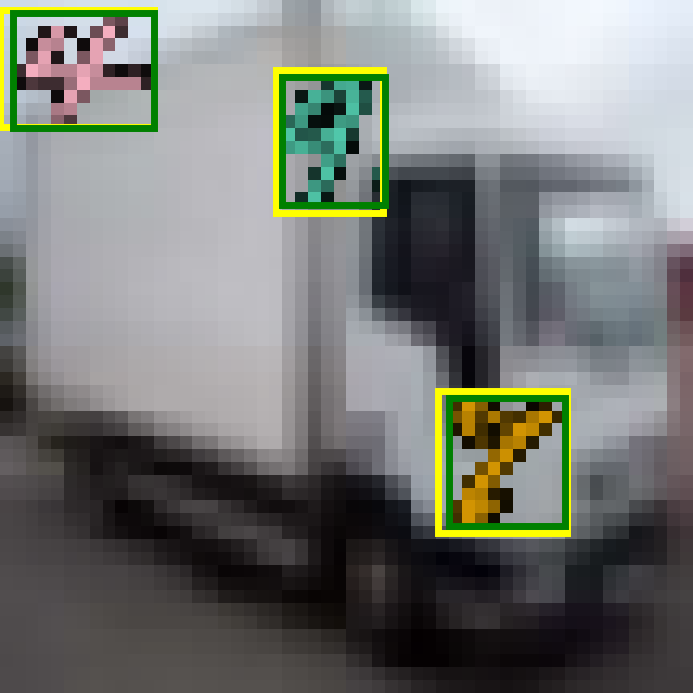}
\includegraphics[width=0.12\textwidth]{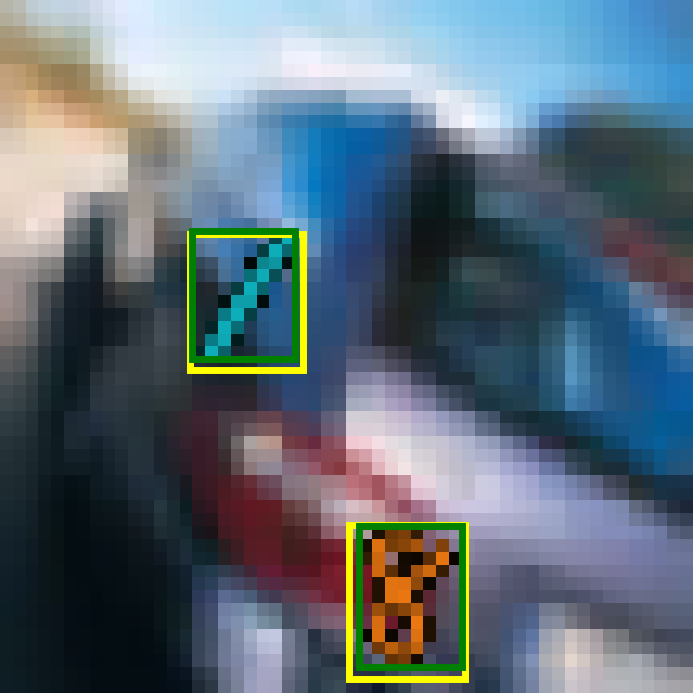}
\includegraphics[width=0.12\textwidth]{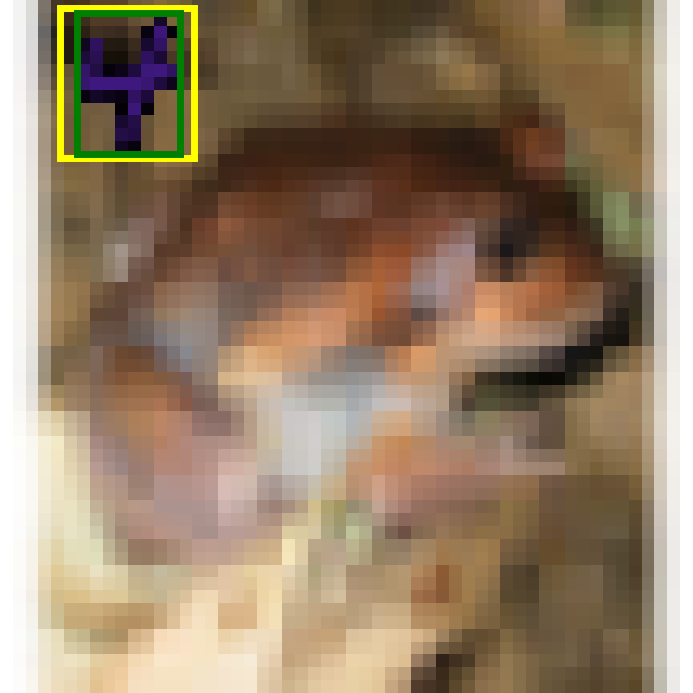}
}
\vline
\subfloat[Multi-MNIST-ImageNet]{
\includegraphics[width=0.12\textwidth]{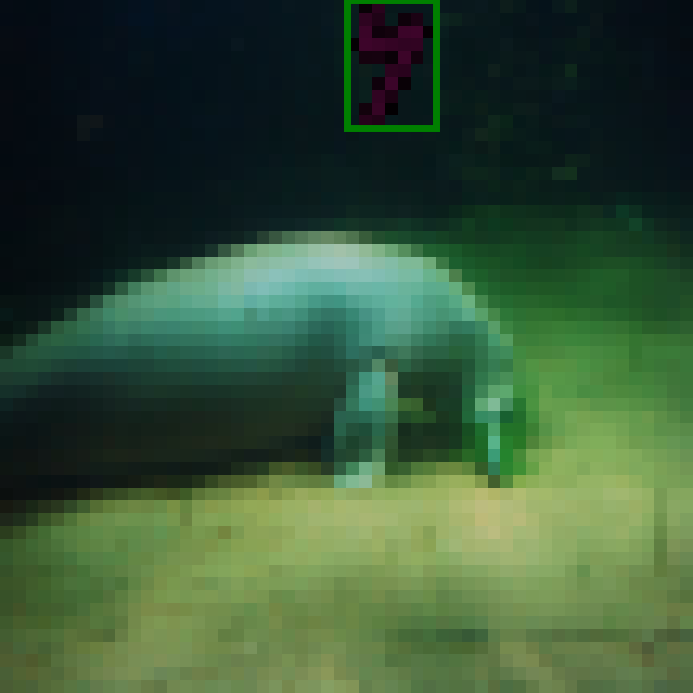}
\includegraphics[width=0.12\textwidth]{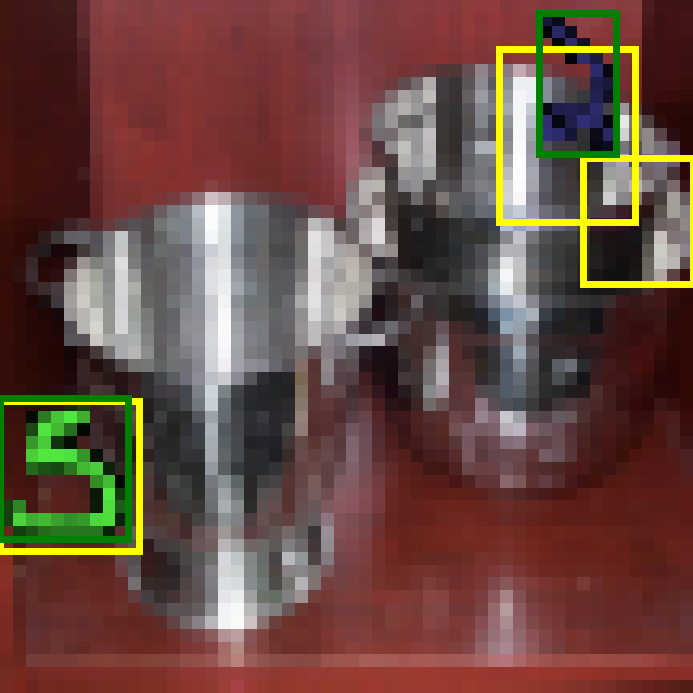}
\includegraphics[width=0.12\textwidth]{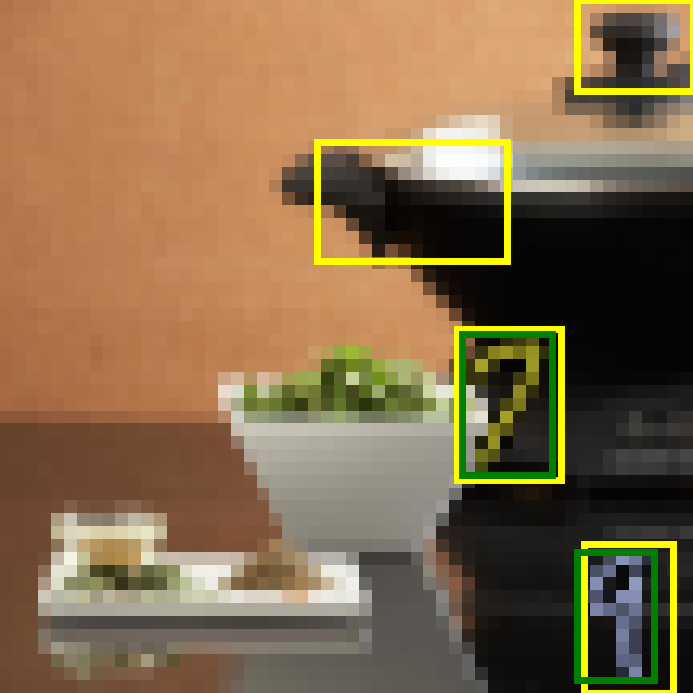}
\includegraphics[width=0.12\textwidth]{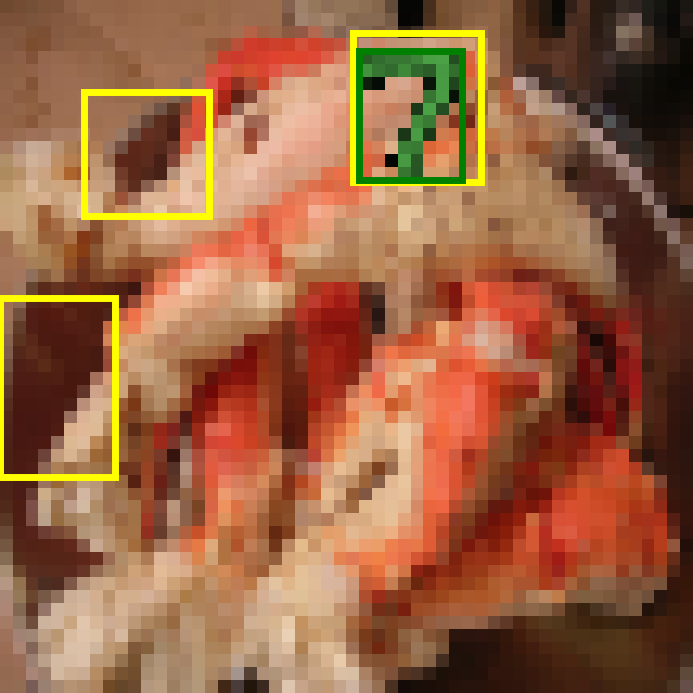}
}
\caption{Examples of detection results in \textit{ascending order w.r.t their data parameters $\sigma_i$}. GT bounding boxes are annotated in \textcolor{green}{green} and proposed boxes are in \textcolor{yellow}{yellow}. 
The leftmost image has the smallest data parameter and is down-weighted for learning, since the only digit is missing.
The rightmost image has the largest data parameter and is easiest to learn, since it contains only one digit.
}
\label{fig:dp-example}
\end{figure}

\paragraph{Task Description}
We first introduce a single-step task called Multi-MNIST.
Given a 54x54 image on which some MNIST digtis are scattered randomly, the task is to calculate the sum of the digits. The training set consists of 60000 images and each image has 1 to 3 MNIST digits, while the the test set consists of 10000 images with 4 MNIST digits. The task is inspired by AIR~\cite{eslami2016attend}, but we render digits with random colors on complex backgrounds from two different sources: CIFAR-10~\cite{krizhevsky2009learning} and ImageNet~\cite{imagenet_cvpr09}.
The datasets generated are denoted by Multi-MNIST-CIFAR and Multi-MNIST-ImageNet, respectively. Examples images are shown in Fig~\ref{fig:dp-example}.

\paragraph{Method Details} 
In this task, we train all the baselines in a supervised learning manner, but it is equivalent to train a policy by REINFORCE~\cite{sutton_book} with appropriate rewards.
Since this task is irrelevant to the relationship between objects, we use an empty graph (without edges) as the object-centric graph. The node input is a patch cropped from the image according to the bounding box of the corresponding object and resized to 16x16.
Then we use a CNN to encode node features, and apply a global-add-pooling to readout a global feature over all the nodes, followed by an MLP to predict the summation. And the policy GNN is implemented as PointNet~\cite{qi2017pointnet}.

We compare our method with two baselines: plain CNN and Relation Net~\cite{zambaldi2018deep}.
For the two baselines, we flatten the feature map generated by the convolution layers and apply an MLP to predict the summation. It seems to be the best design choice according to our preliminary attempts. Other design choices are discussed in the supplementary.
Besides, we apply the \emph{RandomResizeCrop} to images as the data augmentation for all baselines and our model.

\paragraph{Results}
Table~\ref{tab:multi-mnist-quantitative} shows the quantitative comparison between our GNN-based policy and two baselines, as well as ablation studies.
To measure the performance of the task, we report the accuracy, where the prediction is correct if its absolute difference with the ground truth is less than 0.5.
The recall/AP@0.25 of the object detectors on Multi-MNIST-CIFAR and Multi-MNIST-ImageNet are 93.7/92.4 and 93.8/31.5 respectively.
Our GNN-based policy outperforms two baselines by a large margin, which shows the advantage of our method over CNN-based method w.r.t compositional generalization.
Our policy GNN trained with ground truth boxes achieves high accuracy (81.4 and 69.8), which verifies our implementation.
However, it is still imperfect, which implies the inherent difficulty of this task.
Besides, our policy GNN fails to generalize well without data parameters.
The effect of data parameters is analyzed in Fig~\ref{fig:dp-example}.

\begin{table}[t]
\centering
\small
\setlength{\tabcolsep}{2pt}
\renewcommand{\arraystretch}{1.1}
\subfloat[Multi-MNIST-CIFAR]{
    \begin{tabular}{lll}
    \toprule
    Method                     & Train Acc & Test Acc \\
    \midrule
    CNN                  & 92.0(1.7)          & 30.7(4.9)         \\
    Relation Net             & 97.4(0.5)           & 18.2(13.3)          \\
    GNN+SPACE (ours)                 & 83.1(0.1)          & \textbf{51.2(3.8)}         \\ 
    \midrule
    GNN+GT boxes & 99.5(0.1)           & 81.4(2.0)         \\
    GNN+SPACE (w/o DP)        & 86.1(0.6)           & 29.1(3.6)         \\
    CNN (with DP)        & 86.9(1.2)           & 49.8(2.7)         \\
    \bottomrule
    \end{tabular}
}
\hfill
\subfloat[Multi-MNIST-ImageNet]{
    \begin{tabular}{lll}
    \toprule
    Method                     & Train Acc & Test Acc \\ 
    \midrule
    CNN                 &  90.5(2.9)           & 12.0(2.1)         \\
    Relation Net        &  96.4(0.8)         & 8.4(4.7)         \\
    GNN+SPACE (ours)    &  80.2(0.2)         & \textbf{51.2(1.2)}         \\ 
    \midrule
    GNN+GT boxes &  99.1(0.1)          & 69.8(10.5)         \\
    GNN+SPACE (w/o DP)        & 80.4(0.6)          & 27.7(1.3)         \\
    CNN (with DP)        & 83.7(2.7)           & 14.2(2.1)         \\
    \bottomrule
    \end{tabular}
}
\vspace{2mm}
\caption{Quantitative results on Multi-MNIST. The average with the standard deviation (in the parentheses) over 5 trials is reported. \emph{DP} stands for data parameters introduced in Sec~\ref{sec:gnn_learning}.}
\vspace{-2mm}
\label{tab:multi-mnist-quantitative}
\end{table}

\begin{figure}[h]
\centering
\subfloat[CIFAR-Recon]{
\includegraphics[width=0.24\textwidth]{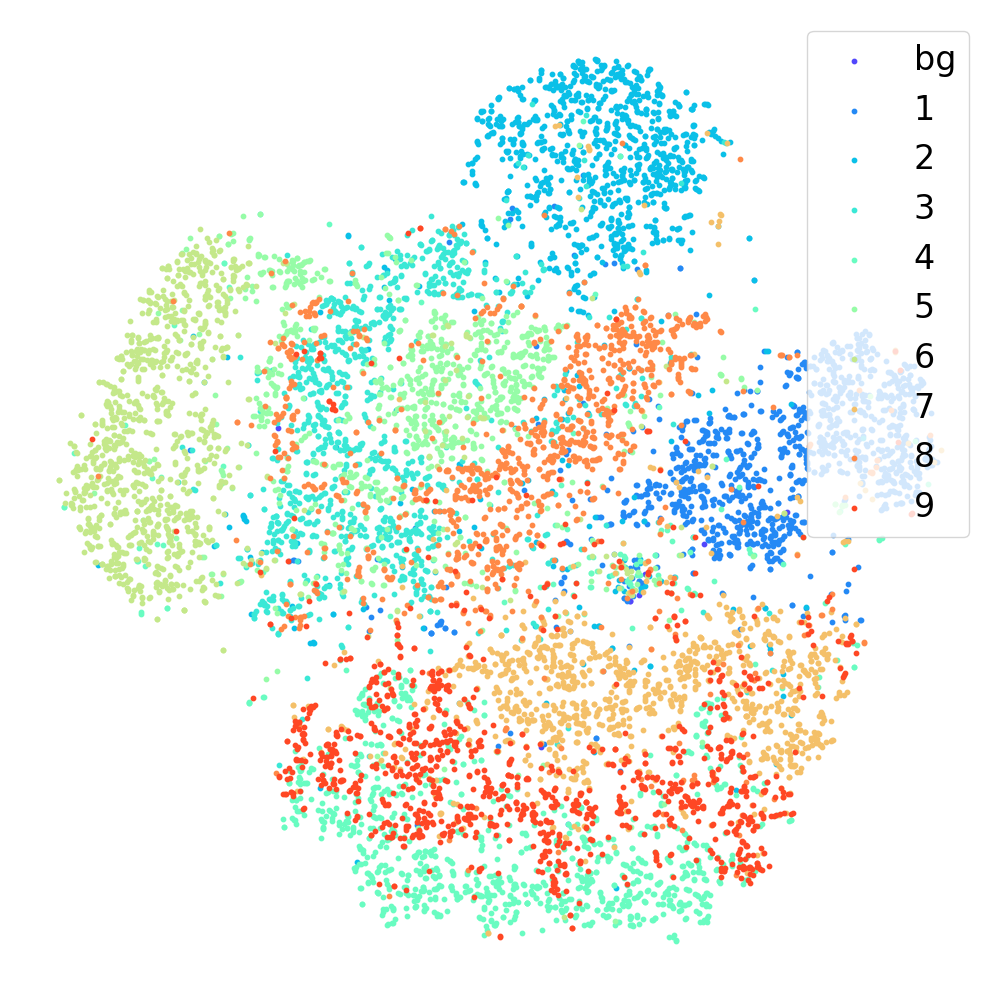}
}
\subfloat[CIFAR-Task]{
\includegraphics[width=0.24\textwidth]{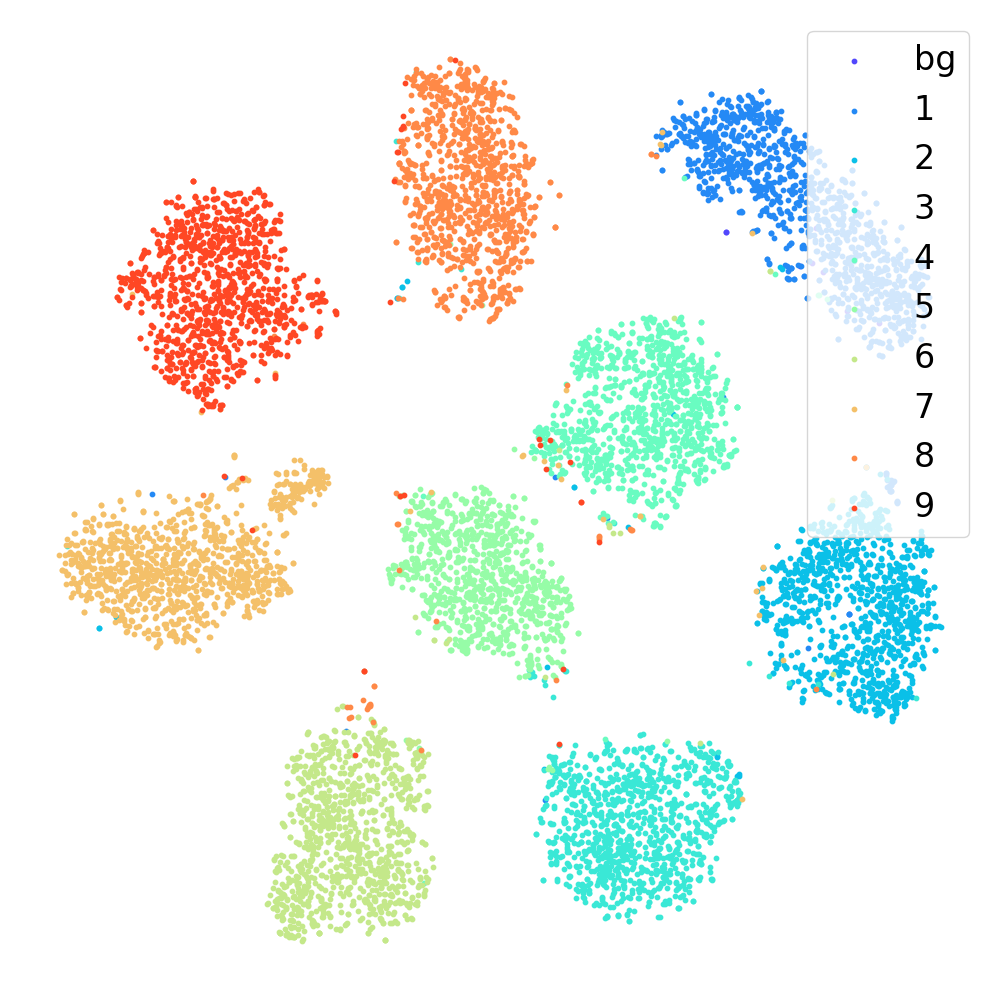}
}
\subfloat[ImageNet-Recon]{
\includegraphics[width=0.24\textwidth]{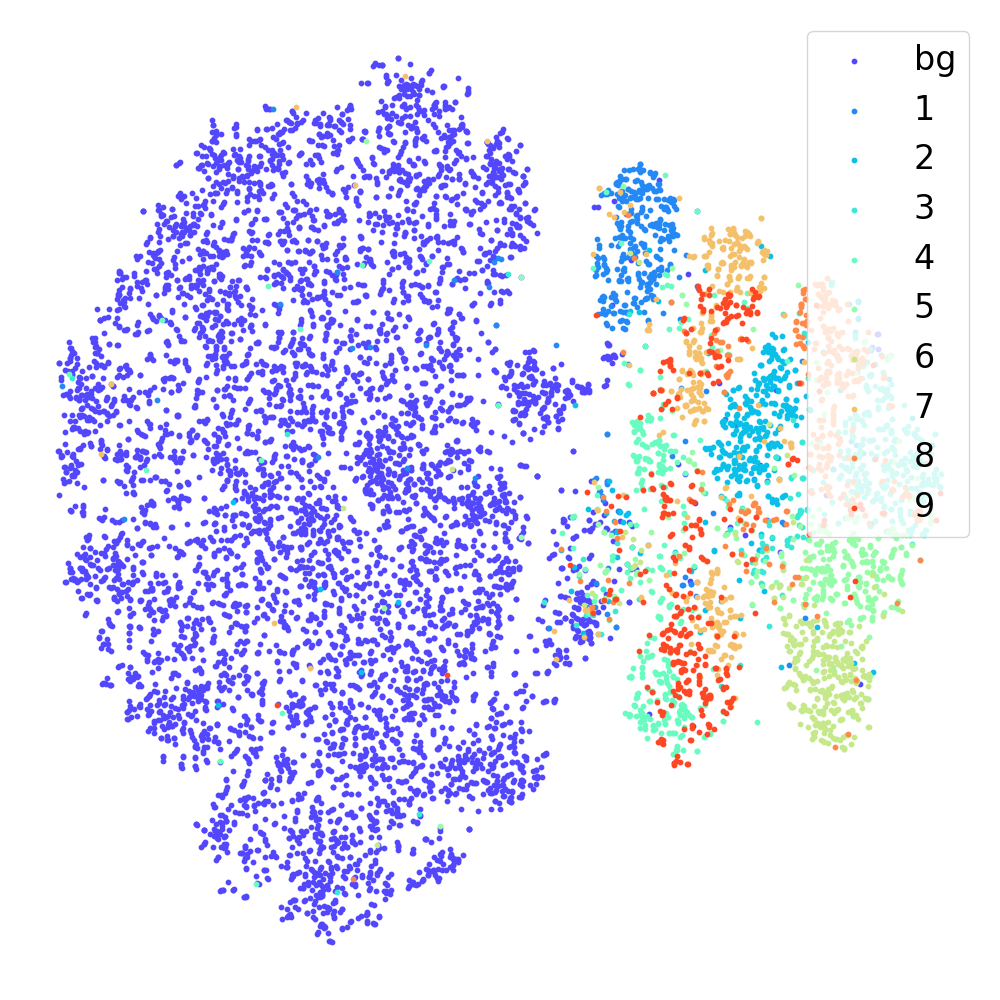}
}
\subfloat[ImageNet-Task]{
\includegraphics[width=0.24\textwidth]{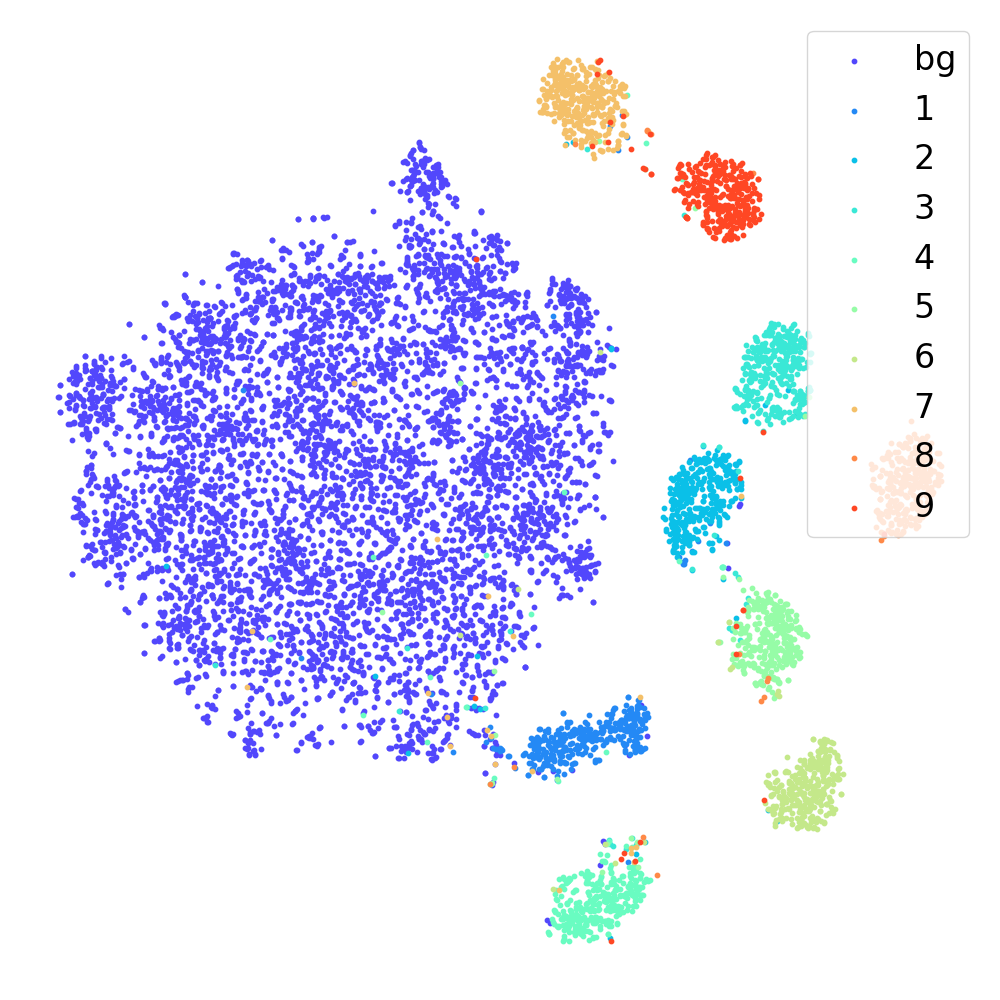}
}
\caption{t-SNE~\cite{maaten2008visualizing} visualization of the learned object features by SPACE~\cite{lin2020space} and our policy GNN.
For CIFAR-Recon, the object features are the latent features used in SPACE~\cite{lin2020space} to reconstruct foreground objects on Multi-MNIST-CIFAR.
For CIFAR-Task, the object features are node features encoded by CNN used in our policy GNN to solve the task on Multi-MNIST-CIFAR.
}
\label{fig:tsne-multi-mnist}
\end{figure}

\paragraph{Analysis}
One advantage of our framework is that it supports interpretable model diagnosis, thanks to the modular design and explicit object modeling.
First, object attributes emerge when the learned object features are clustered.
Fig~\ref{fig:tsne-multi-mnist} visualizes the learned object features by t-SNE~\cite{maaten2008visualizing}.
For each detected object, we label it with the class of its closet ground truth object if the overlap is larger than 0.25, otherwise it is labeled as background.
It is observed that task-driven object features are more distinguishable compared to reconstruction-driven ones.

\subsection{FallingDigit}
\label{sec:exp-falling-digit}
\begin{figure}[h]
\centering
\subfloat[FallingDigit-Black]{
\includegraphics[width=0.15\textwidth]{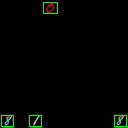}
\includegraphics[width=0.15\textwidth]{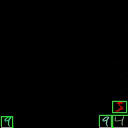}
\includegraphics[width=0.15\textwidth]{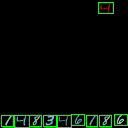}
}
\hfill
\subfloat[FallingDigit-CIFAR]{
\includegraphics[width=0.15\textwidth]{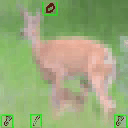}
\includegraphics[width=0.15\textwidth]{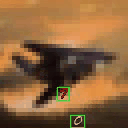}
\includegraphics[width=0.15\textwidth]{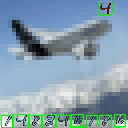}
}
\caption{
Examples of FallingDigit games with different backgrounds.
Object proposal generated by our improved SPACE are annotated in \textcolor{green}{green} bounding boxes.
}
\label{fig:falling-digit-example}
\end{figure}

\paragraph{Task Description}
Inspired by the classical video game ``Tetris'', we design this FallingDigit game. In this game,  there is a digit falling from the top and some target digits lying on the bottom. The player needs to control the falling digit to hit the closest target digit (in terms of digit value). At each step, the falling digit can be moved to one of the three direction: down-left, down, down-right. The player receives +1 reward when the falling digit hits the correct target, and the target digit will be cleared. The player receives -1 reward when it hits the bottom or wrong targets, and the wrong targets will not be cleared. The falling digit will disappear when it hit bottom or any target digits, and a new falling digit will be created at the next timestep. The positions of all digits are random.
An episode will terminate when all the target digits are cleared or the player has taken 100 actions. The training environment has 3 target digits, while the agent is tested in the environments with more target digits. To generalize, the agent has to localize and identify numbers from pictures, learn to compare the difference between digits, and handle interactions between distant pixels. We create two variants of this game: FallingDigit-Black (background is black) and FallingDigit-CIFAR (background is selected from a subset of CIFAR). Digits are from MNIST. Examples are shown in Fig~\ref{fig:falling-digit-example}.

\paragraph{Method Details}
For our framework, we first train a teacher policy by DQN~\cite{mnih2015human} in the training environment, which can converge to a near-optimal solution. 
The architecture of teacher policy is Relation Net~\cite{zambaldi2018deep}.
The teacher policy is used to collect 60000 images through the interaction with the environment, and label them with the taken actions, as illustrated in Sec~\ref{sec:demonstration-acquisition}. When building the demonstration dataset, we filter out a few episodes with low rewards so as to avoid providing incorrect demonstrations. Then, we train the self-supervised object detector on the collected images, and train the policy GNN based on the generated object proposals. Since it is critical to reason about the relationship between the falling digit and other target digits in this game, we use a complete graph as the object-centric graph. The node input includes the bounding box position and a patch cropped from the image according to the bounding box, which is resized to $16\times 16$. The policy GNN is implemented as EdgeConv~\cite{dgcnn}.
We compare our method with two baselines: plain CNN and Relation Net~\cite{zambaldi2018deep}, both of them are trained by DQN algorithm.
To make a fair comparison, the RL agents are trained on a fixed set of episodes, which are also the source of the demostration dataset.

\begin{figure}[h]
\centering
\captionsetup[subfigure]{labelformat=empty}
\subfloat{\includegraphics[width=0.45\textwidth]{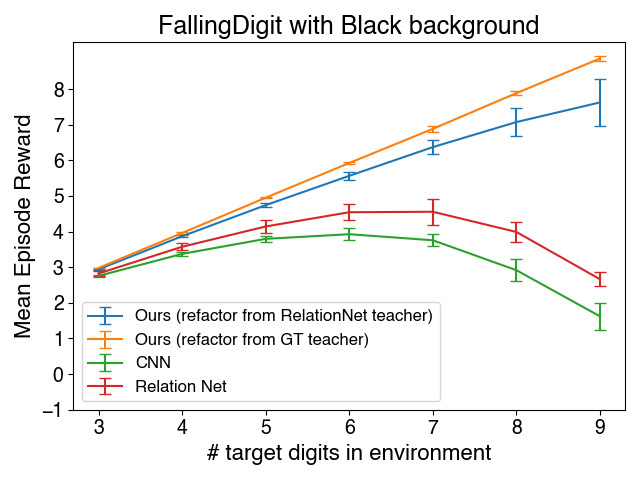}}
\hspace{4 mm}
\subfloat{\includegraphics[width=0.45\textwidth]{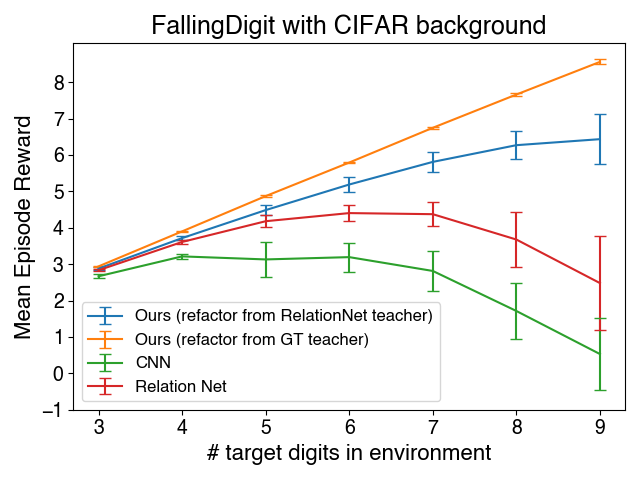}}
\caption{Quantitative results of generalizing to the FallingDigit environments with different number of target digits. Ours uses GNN+SPACE architecture. CNN and Relation Net are trained by DQN. For ours (refactor from Relation Net teacher), we report the mean episode reward over 9 different runs (3 different RL teacher runs and 3 refactorization runs per teacher). For the other baselines, the result is averaged over 3 runs. The error bar shows standard deviation.}
\label{fig:falling-digit-generalization}
\end{figure}

\paragraph{Results}
Fig~\ref{fig:falling-digit-generalization} shows the mean episode rewards of different methods in the environment with different number of target digits.
In all the environments, our agent outperforms the Relation Net (which is our RL teacher) and CNN baselines. Our agent can generalize well to the environments with 9 target digits, while the performance of Relation Net and CNN gradually decrease when the number of objects is increasing. We also generate the ground truth action labels by a heuristic program, which can serve as the teacher policy.
It results in even better generalizability than the RL teacher, since it provides perfect and more consistent demonstration.

\subsection{BigFish}
\label{sec:exp-bigfish}
\begin{figure}[h]
\centering
\begin{minipage}{.6\textwidth}
    \centering
    \includegraphics[width=0.3\linewidth]{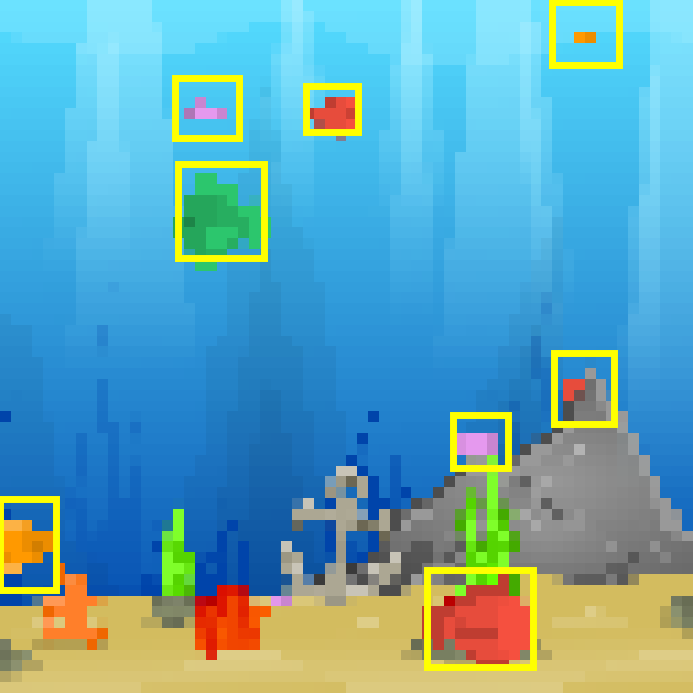}
    \hspace{7 mm}
    \includegraphics[width=0.3\linewidth]{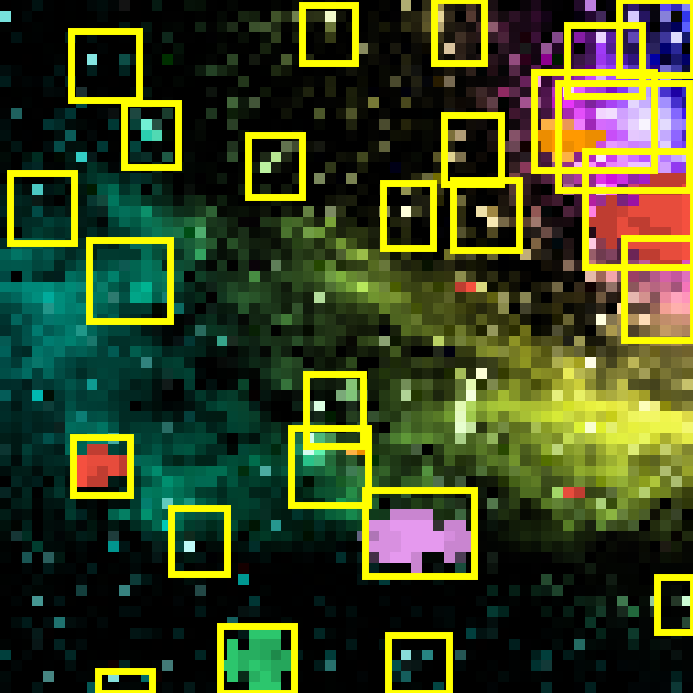}
    \caption{Examples of BigFish games and object proposals generated by SPACE (annotated in \textcolor{yellow}{yellow} boxes.) The left one is sampled from training environments and the right one is from test environments.}
    \label{fig:bigfish-example}
\end{minipage}%
\hspace{4 mm}
\begin{minipage}{.3\textwidth}
    \centering
    \includegraphics[width=0.6\linewidth]{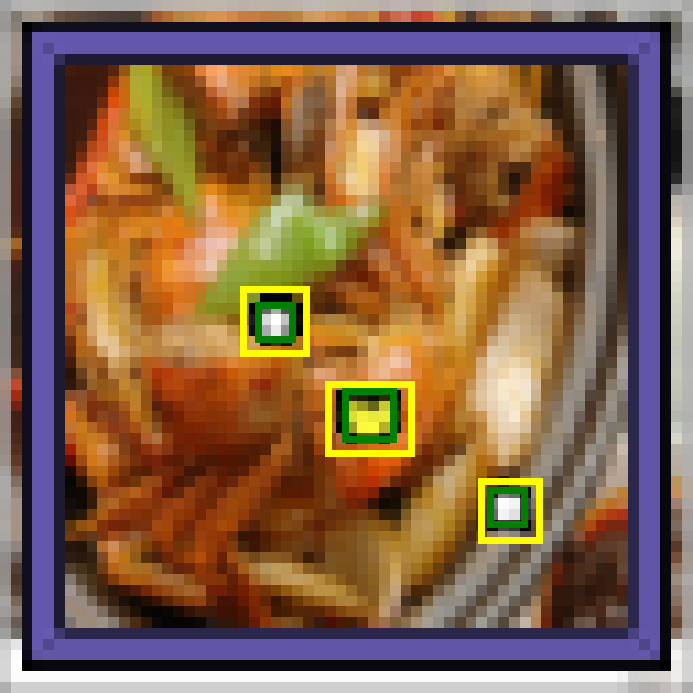}
    \caption{Examples of our customized Pacman game.
    GT bounding boxes are in \textcolor{green}{green}, object proposals are in \textcolor{yellow}{yellow}.
    }
    \label{fig:pacman-example}
\end{minipage}
\end{figure}

\paragraph{Task Description}
BigFish is a game from ProcGen Benchmark~\cite{cobbe2019procgen}, which provides a challenging generalization scenario to RL agents.
Fig~\ref{fig:bigfish-example} shows some examples of the game.
In this game, the player needs to eat fish smaller than itself to gain rewards and make itself larger. Contacting with a larger fish will terminate the episode. 
To evaluate the compositional generalizability w.r.t different backgrounds, we modify the BigFish game to create new test environments with unseen complicated backgrounds.
In our experiment, we train agents on level 0-199 and zero-shot test on level 500-599 with unseen backgrounds. The difficulty mode of the game is set to easy.

\paragraph{Method Details}
Following the baseline provided by Procgen Benchmark \cite{cobbe2019procgen}, we use PPO\cite{schulman2017proximal} to train a CNN-based policy network.
When collecting the demonstration dataset, we use $\epsilon$-greedy exploration strategy to increase the diversity of states. And we apply data augmentation by adding some object proposals with low confidence scores. The details can be found in the supplementary.
Each image in the demonstration dataset is labeled with the softmax logits of action by the teacher policy.
Since the task requires reasoning about the relation between the player and other fishes, we also use a complete graph. The architecture details of our policy GNN are in supplementary. We compare our method against CNN and Relation Net, both are trained by PPO for 200M frames. And the CNN serves as the teacher in our framework.

\paragraph{Results}
Table \ref{tab:bigfish_gen} shows that our refactorized GNN-based policy outperforms the two RL baselines in the test environments with unseen complicated backgrounds, which shows the object-centric graph as inductive bias makes the policy more robust to background changes. In the training environment, the CNN gets 29.65 and the Relation Net gets 28.22. Since our GNN-based policy is trained with augmented low-confidence object proposals, it performs slightly worse than the CNN teacher in the training environment (gets 24.86).

\begin{table}[h]
\centering
\begin{tabular}{lc}
\toprule
Method               & Test on unseen backgrounds \\
\midrule
CNN                &  4.40(1.90)     \\
Relation Net       &  4.54(1.22)        \\
Ours (refactor from CNN)       &  \textbf{6.05(2.44)}        \\
\bottomrule
\end{tabular}
\vspace{2mm}
\caption{Quantitative results on BigFish. Ours uses GNN+SPACE architecture. CNN and Relation Net are trained by PPO. For ours, we report the mean episode reward over 24 different runs (4 different RL teacher runs, 3 different demonstration datasets per teacher and 2 refactorization runs per dataset). For the other baselines, the result is averaged over 4 runs. The standard deviation is in the parentheses.}
\vspace{-2mm}
\label{tab:bigfish_gen}
\end{table}

\subsection{Pacman}

\label{sec:exp-pacman}
\paragraph{Task Description}
We build a customized Pacman game based on \cite{berkeley_pacman}. Fig~\ref{fig:pacman-example}  shows an example. The objective of this game is to control the Pacman to eat all the dots.
The initial positions of all objects are random at each episode.
At each time step, the agent can take one of the four actions, to move one step towards left, right, up or down in a 14x14 grid.
The agent receives +1 point as the reward for eating a dot, and -0.01 point for each move. An episode will terminate when all the dots are eaten or the agent takes 100 actions. The training environment is initialized with 2 dots, while the agent is tested in the environments initialized with more dots. The backgrounds in this game are random selected from a subset of ImageNet.

\paragraph{Method Details}
The basic experimental method is similar to FallingDigit in Sec~\ref{sec:exp-falling-digit}.
In this experiment, we focus on comparing the different student network architectures. We use a CNN-based policy as the teacher since it achieves almost perfect scores in the training environments. Then we refactor it into obejct-centric GNN, CNN and Relation Net. The architecture details of each student can be found in supplementary materials.

\begin{table}[h]
\centering
\setlength{\tabcolsep}{1em}
\begin{tabular}{lllll}
\toprule
Method                          & 2 & 3 & 5 & 10 \\
\midrule
CNN (RL)                 & 1.86(0.02)  & 2.63(0.14)  & 1.64(0.31)  & -0.41(0.09)   \\
Relation Net (RL)        & 1.84(0.04)  & 2.76(0.08)  & 4.31(0.16)  & 4.40(0.32)   \\
CNN (refactor)           & 1.86(0.00)  & 2.80(0.00)  & 4.70(0.01)  & \textbf{9.17(0.09)} \\
Relation Net (refactor)  & 1.86(0.00)  & 2.80(0.01)  & 4.70(0.00)  & \textbf{9.11(0.23)} \\
GNN+SPACE (refactor)     & 1.86(0.02)  & 2.80(0.02)  & 4.63(0.08)  & \textbf{8.32(0.31)} \\
\bottomrule
\end{tabular}
\vspace{2mm}
\caption{Quantitative results on Pacman. The mean episode reward with the standard deviation (in the parentheses) over three different runs is reported. \emph{RL} and \emph{refactor} indicates the ways we train this architecture.}
\vspace{-2mm}
\label{tab:pacman}
\end{table}

\paragraph{Results}
Table~\ref{tab:pacman} shows the mean episode rewards of different methods in the environments with different number of dots. The refactorized GNN policy generalizes well to the environments with 10 dots, and outperforms all RL baselines. However, in this Pacman environment, the refacorized CNN and refactorized Relation Net also generalizes very well to the environments with 10 dots. 
Note that in the three other environments mentioned above, the GNN student performs better than CNN and Relation Net students according to our experiments (see project website for more results). This shows that, for some environments like Pacman, a generalizable student policy does not have to be object-centric GNNs. We feel that the main reason for the effectiveness of CNN and Relation Net in this environment is because it could make reasonable decisions just by looking at nearby regions. Another observation is that the refactorized CNN student generalizes better than its CNN teacher, but this phenomenon is not observed in other three environments.

\section{Related Work}
\paragraph{Structured RL}
Many works have investigated structured representations and structured policies in the RL literature.
\cite{kansky2017schema,bapst2019structured, davidson2020investigating, garnelo2016towards} show that structured policies induced by structured representations(\eg, symbols, objects and scene graphs) generalize better than unstructured counterparts.
Other works explicitly model the structure of an agent, and use GNN to learn either a policy~\cite{wang2018nervenet} or a forward model~\cite{sanchez2018graph}.
However, they rely on ground truth object(part) information or graph structures.
Another series of works~\cite{zambaldi2018deep,mott2019towards} make use of attention mechanism to augment model-free deep RL agents, which improves performance, learning efficiency, generalization and interpretability.
However, attention maps can not provide as interpretable analysis as our framework.

\cite{wang2019deep} proposes an object-centric perception approach to deep control problems, which shows better generalizability and interpretability.
Our method differs from it in several aspects:
1) our demonstration dataset is generated by RL algorithms while their dataset is annotated by human;
2) they do not include GNN to reason about objects;
3) our object detector is unsupervised while their object detector is trained with supervision.

\paragraph{Self-Supervised Object Discovery}
Our object detector is built upon a self-supervised object detector, SPACE~\cite{lin2020space}.
It extends \cite{eslami2016attend,crawford2019spatially} and decomposes an image into salient foreground objects and background stuff, self-supervised by reconstruction.

Several works attempt to combine self-supervised scene decomposition approaches with model-based RL.
COBRA~\cite{watters2019cobra} employs MONet~\cite{burgess2019monet} to obtain object latent representations, which are used to train an exploration policy.
OP3 \cite{veerapaneni2019entity} extends IODINE~\cite{greff2019multi} and applies a planning module on top of learned entities. Their objects are represented by latent slots.  Both work show the ability to generalize to novel tasks; however, 
they suffer from the drawback of scene decomposition approaches and are not able to handle multiple instances of the same category (shape and color), thus cannot be used to solve our experiment environments.
Other approaches rely on motion clues.
\cite{goel2018unsupervised} learns moving object segmentation in an unsupervised fashion, which helps improve sample efficiency.
\cite{du2019task} makes use of a video prediction model capable of capturing object dynamics to achieve faster convergence and better generalization.
\cite{kulkarni2019unsupervised} learns unsupervised keypoint detection, and uses both the keypoint co-ordinates and corresponding image features to improve sample efficiency.
Different from our method, they augment CNNs with either features or outputs from their perception modules, rather than directly learn policies from detections.

\section{Discussion and Future Work}
Our two-stage scheme allows us to decouple the efforts to address the optimization challenge and the generalization challenge. We observe that, with our SPACE empowered object-centric GNNs, we can refactor a high reward policy learned in the training environments into a student policy with better generalizability. Practically, this scheme also improves training speed, \eg, training a GNN by RL with online-generated object proposals is super slow, but training a CNN teacher and refactoring it into a GNN is faster. While our experiments have been focusing on compositional generalizability, we feel that this scheme may benefit other types of generalizabilities, like variation in visual appearance and reward function, as well. 

Besides the overall framework, we also improved the SPACE method, which learns a self-supervised object detector to be used by GNN. Adding the object proposal module makes the framework closer to a white-box system. Along with the learning process, we can also discover task-relevant objects and attributes. The human readable representation, including the objects and attributes, can benefit diagnosis of the algorithm, result interpretation, and knowledge increment for transfer learning.  

Some of the design choices in this work are a double-edged sword: 1) We assume the accessibility of a high-recall object proposal algorithm. Compared with more end-to-end methods, such as plain CNN-based RL~\cite{mnih2015human} or attention-based relational reasoning framework~\cite{zambaldi2018deep}, the object proposal algorithm serves as a strong inductive bias, which may fail the policy learning when the inductive bias is inappropriate; 2) We rely on the Graph Neural Network (GNN) to achieve compositional generalizability. The limitations of the GNN toolkit would also restrict the power of our approach. Nonetheless, the research into more powerful GNNs is a hot topic and the technique is improving; 3) Our policy is object-centric but has ignored the important role stuffs play for reasoning. 

Finally, this work has left many open questions, even if we have explored to the best of our endeavours. First, while we are aware that policy consistency and learnability are important, we lack rigor definition and analysis to this issue. Second, our understanding to how the policy refactorization would work is quite limited. A particularly interesting case is that, policy refactorization tends to change the generalizability of a teacher policy, even if the student policy uses exactly the same architecture (\eg, CNN in Pacman could generalize much better as a student than as a teacher, but RelationNet in FallingDigit generalized slightly worse as a student than as a teacher). We conjecture it is relevant to the property of the network, the game, and the order that data are fed. We leave all these open questions to future work.

\newpage
\section{Acknowledgement}
This research was supported by NSF grant IIS-1764078. We also acknowledge Sirui Xu and Shuang Liu for their help in discussions and experiments.
\section{Broader Impact}
Our work is a basic step towards building autonomous agents that can train in limited environment and perform well in more complicated environment with similar reasoning rationale, using vision as the primary information source. Particularly, we try to build a system that makes decisions in a way that human may interpret. From an ethical aspect, it is helpful to build AI that humans can better communicate with.
\bibliography{references}
\bibliographystyle{plain}

\end{document}


\maketitle

\section{Overview}
This supplementary material includes implementation details relevant to network architectures and hyperparameters, as well as additional experiments to analyze the robustness of our two-stage framework.
Sec~\ref{sec:space} illustrates our improvements made to SPACE~\cite{lin2020space}.
Sec~\ref{sec:exp-multi-mnist-supp},~\ref{sec:exp-falling-digit-supp},~\ref{sec:exp-bigfish-supp}, and \ref{sec:exp-pacman-supp} describe the network architectures and hyperparameters used in our experiments on Multi-MNIST, FallingDigit, BigFish, and Pacman, respectively.
A robustness analysis is presented in Sec~\ref{sec:robustness-analysis}, where we also show the advantages of our two-stage framework over end-to-end training object-centric GNNs by RL.

\section{Improvements of SPACE}
\label{sec:space}
In this section, we provide a recap of SPACE~\cite{lin2020space} along with the improvements.
We refer readers to the original paper for more complete explanation.
SPACE is a unified probabilistic generative model that combines both the spatial-attention (foreground) and scene-mixture (background) models.
It assumes that an image (or scene) can be decomposed into foreground and background latents: $z^{fg}$ and $z^{bg}$.
The foreground $z^{fg}$ consists of a set of independent foreground objects $z^{fg}=\{z^{fg}_i\}_{i=1}^N$.
The observed image $x$ is modeled as a sample from the pixel-wise Gaussian mixture model, which is a combination of foreground and background image distributions, as illustrated in Eq~\ref{eq:generation}.
\begin{equation}
    p_{\theta}(x|z^{fg}, z^{bg}) = \alpha p_{\theta}(x|z^{fg}) + (1 - \alpha) p_{\theta}(x|z^{bg}) 
    \label{eq:generation}
\end{equation}
, where $\theta$ is the parameters of the generative network, $\alpha$ is the foreground mixing probability.

\paragraph{Foreground}
The foreground image component is modeled as a Gaussian distribution $p(x|z^{fg}) \sim \mathcal{N}(\mu^{fg}, \sigma^{fg})$.
It is represented as structured latents.
Concretely, the image is divided into $H \times W$ cells, and each cell represents a potential foreground object.
Each cell is associated with a set of latent variables $\{z^{pres}_i, z^{where}_i, z^{depth}_i, z^{what}_i\}$.
The underlying idea is similar to RCNN~\cite{ren2015faster} and YOLO~\cite{redmon2016you}.
$z^{pres} \in \mathcal{R}$ is a binary random variable indicating the presence of the object in the cell.
$z^{what}$ models the object appearance and mask, and $z^{depth} \in \mathcal{R}$ indicates the relative depth of the object.
$z^{where} \in \mathcal{R}^4$ parameterizes the object bounding box.
For each cell that $z^{pres}_i=1$, SPACE uses $z^{what}_i$ to decode the object reconstruction and its mask.
The object reconstruction is positioned on a full-resolution canvas using $z^{where}_i$ via the Spatial Transformer Network~\cite{jaderberg2015spatial}.
The object depth $z^{depth}_i$ is then used to fuse all the object reconstructions into a single foreground image reconstruction $\mu^{fg}$.
In practice, $\sigma^{fg}$ is treated as a hyperparameter.

\paragraph{Improvement: Bounding Box Parameterization}
As illustrated in \cite{github2020space}, SPACE~\cite{lin2020space} is somehow unstable to train and sensitive to some hyperparameters, \eg, prior object sizes.
In the original paper, $z^{where}$ is decomposed into two latent logits $z^{shift}$ and $z^{scale}$, activated by $tanh$ and $sigmoid$, which represent the shift and scale of the bounding box respectively.
Thus, to change the prior size of objects, it is required to tune those non-intuitive logits.
Besides, the scale $z^{scale}$ is relative to the whole image instead of the cell, which makes the model more fragile.

In our implementation, following R-CNN~\cite{ren2015faster}, we reparameterize the bounding box $(x_{ctr}, y_{ctr}, w, h)$ as an offset $(dx, dy, dw, dh)$ relative to a pre-defined anchor $(x_a, y_a, w_a, h_a)$ centered at the cell.
For simplification, we associate each cell with only one anchor.
It can be easily extended to multiple anchors per cell~\cite{ren2015faster} or multiple levels~\cite{lin2017feature}.
Eq~\ref{eq:bounding-box-regression} illustrates the formula of the `bounding-box regression' reparameterization.
\begin{equation}
    \begin{split}
    dx &= (x_{ctr} - x_a) / w_a \\
    dy &= (y_{ctr} - y_a) / h_a \\
    dw &= \log((w - w_a) / w_a) \\
    dh &= \log((h - h_a) / h_a) \\
    \end{split}
    \label{eq:bounding-box-regression}
\end{equation}
Compared to the original implementation, ours supports more intuitive interpretation of the hyperparameters related to the bounding box.
The priors of $(dx, dy, dw, dh)$ can be simply modeled as zero-mean Gaussian distributions.
And the variances of Gaussian distributions can be tuned to control how much the bounding box can be different from the anchor.

\paragraph{Improvement: Background}
In the original paper~\cite{lin2020space}, the background is modeled as a sequence of segments by GENESIS~\cite{engelcke2019genesis}.
From our experiments, we find that the key to design a background module is its capacity.
Thus, it is not necessary to use complex and expensive models, \eg, MONet~\cite{burgess2019monet}, GENESIS~\cite{engelcke2019genesis}, IODINE~\cite{greff2019multi}, especially for the RL applications.
It is even not necessary to use a VAE.
By using a much simpler background module, our improved SPACE model can be easily trained with a single GPU and a higher, unified learning rate, which results in faster convergence.
In contrast, the default configuration of SPACE requires 4 GPUs to train, separate learning rates for different modules, and other sophisticated tricks (fixed $\alpha$ at the beginning of training).

\paragraph{Inference and Training}
All the random variables follow Gaussian distributions, except for $z^{pres}$ follows a Concrete distribution~\cite{maddison2016concrete}.
As the model is a variational autoencoder (VAE), the reparameterization trick~\cite{kingma2013auto} is used to optimize the ELBO.

\section{Multi-MNIST}
\label{sec:exp-multi-mnist-supp}

\subsection{Self-supervised Object Detector}
SPACE~\cite{lin2020space} consists of several modules: foreground image encoder, glimpse encoder, glimpse decoder, background image encoder, background image decoder.
We refer readers to \cite{lin2020space} for detailed explanation of each module.
For the glimpse of each object, we apply the spatial transformer network~\cite{jaderberg2015spatial} (STN) to crop a patch from the image according to the bounding box and resize it to 14x14.
Table~\ref{tab:arch-space-multi-mnist} shows the architecture of the object detector used in the experiments.
Table~\ref{tab:hyper-space-multi-mnist} shows the hyperparameters of the object detector.

\begin{table}[t]
\centering
\small
\begin{tabular}{llll}
\multicolumn{4}{c}{\textbf{Foreground Image Encoder}} \\
\toprule
Layer       & Resolution   & Stride   & Norm./Act.   \\
\midrule
Input       & 54x54x3      &          &              \\
Conv 3x3    & 54x54x64     & 1        & BN/ReLU      \\
Conv 3x3    & 18x18x64    & 3        & BN/ReLU      \\
Conv 3x3    & 18x18x128    & 1        & BN/ReLU      \\
Conv 3x3    & 6x6x128      & 3        & BN/ReLU      \\
Conv 3x3    & 6x6x256      & 1        & BN/ReLU      \\
Conv 1x1    & 6x6x128      & 1        & BN/ReLU      \\
Conv 1x1    & 6x6x128      & 1        & BN/ReLU      \\
\midrule
\multirow{3}{*}{Conv 1x1}   & 6x6x1 (object presence $z^{pres}$)     & 1     & Sigmoid   \\
& 6x6x4 (bounding box mean $z^{where}$)     & 1     &    \\
& 6x6x4 (bounding box stdev $z^{where}$)     & 1     & Softplus   \\
\bottomrule
\end{tabular}

\subfloat{
\begin{tabular}{lll}
\multicolumn{3}{c}{\textbf{Glimpse Encoder}}   \\
\toprule
Layer                       & Resolution & Norm./Act. \\
\midrule
Input                       & 14x14x3    &            \\
Flatten                     & 588        &            \\
Linear                      & 256        & GN(16)/ReLU    \\
Linear                      & 256        & GN(16)/ReLU    \\
\multirow{2}{*}{Linear}     & 50 (mean $z^{what}$)    \\
                            & 50 (stdev $z^{what}$)    & Softplus  \\
\bottomrule
\end{tabular}
}
\hfill
\subfloat{
\begin{tabular}{lll}
\multicolumn{3}{c}{\textbf{Glimpse Decoder}}   \\
\toprule
Layer                       & Resolution & Norm./Act. \\
\midrule
Input                       & 50         &  \\
Linear                      & 256        & GN(16)/ReLU    \\
Linear                      & 256        & GN(16)/ReLU    \\
Linear                      & 784        & Sigmoid      \\
Reshape                     & 14x14x4    &   \\
\bottomrule
\end{tabular}
}
\\
\subfloat{
\begin{tabular}{llll}
\multicolumn{4}{c}{\textbf{Background Image Encoder}} \\
\toprule
Layer      & Resolution  & Stride  & Norm./Act.  \\
\midrule
Input      & 54x54x3     &         &             \\
Conv 3x3    & 54x54x64     & 1        & BN/ReLU      \\
Conv 3x3    & 18x18x64    & 3        & BN/ReLU      \\
Conv 3x3    & 18x18x128    & 1        & BN/ReLU      \\
Conv 3x3    & 6x6x128      & 3        & BN/ReLU      \\
Conv 3x3    & 6x6x256      & 1        & BN/ReLU      \\
Maxpool    & 256         &         &             \\
\bottomrule
\end{tabular}
}
\hfill
\subfloat{
\begin{tabular}{lll}
\multicolumn{3}{c}{\textbf{Background Image Decoder}}  \\
\toprule
Layer      & Resolution  & Norm./Act.  \\
\midrule
Linear     & 256         & BN/ReLU     \\
Linear     & 256         & BN/ReLU     \\
Linear     & 8478        & Sigmoid      \\
Reshape    & 54x54x3     &     \\
\bottomrule
\end{tabular}
}
\vspace{0.5em}
\caption{
The architecture of the self-supervised object detector for all the experiments on Multi-MNIST.}
\label{tab:arch-space-multi-mnist}
\end{table}

\begin{table}[ht]
\centering
\small
\begin{tabular}{lll}
\toprule
Name                    & Value                           & Schedule    \\
\midrule
max iteration           & 100K                              &             \\
optimizer               & Adam                            &             \\
batch size              & 64        & \\
learning rate           & 1e-3                            &             \\
gradient clip           & 1.0                            &             \\
$z_{pres}$ prior        & $0.1 \to 0.01$       & $10K \to 50K$ \\
$z_{pres}$ temperature  & $2.0 \to 0.1$       &  $10K \to 50K$ \\
$z_{where}$ prior mean  & $0$                 &             \\
$z_{where}$ prior stdev & $0.2$            &             \\
$z_{what}$ prior mean   & $0$                 &             \\
$z_{what}$ prior stdev  & $1.0$            &             \\
$z_{what}$ dimension    & 50            &             \\
$z_{depth}$ prior mean   & $0$                 &             \\
$z_{depth}$ prior stdev  & $1.0$            &             \\
$z_{depth}$ scale       & $10.0$            &             \\
fg recon prior stdev    & $0.15$          &  \\
bg recon prior stdev    & $0.15$          &  \\
\bottomrule
\end{tabular}
\vspace{0.5em}
\caption{The hyperparameters of the self-supervised object detector for all the experiments on Multi-MNIST.}
\label{tab:hyper-space-multi-mnist}
\end{table}

\subsection{Policy GNN}
\paragraph{Network Architectures}
In the Multi-MNIST experiment, the policy GNN is implemented as PointConv~\cite{dgcnn} in Pytorch Geometric~\cite{Fey/Lenssen/2019}. The input graph is an empty graph (without edges). 
Each node corresponds to a detected object and the node feature is an embedded image feature $x_{img}$. We apply the STN to crop a patch from the image according to the bounding box and resize it to 16x16, and then we encode the patch by a CNN to get $x_{img}$. The CNN encoder is denoted by \emph{Image Patch Encoder}.
Note that the bias terms are removed in the fully connected layers after the global readout function.
Table~\ref{tab:arch-gnn-multi-mnist} shows the architecture of GNN and image patch encoder used in Multi-MNIST experiments.

\paragraph{Hyperparameters in Training}
When training the GNN, the batch size is 64.
The initial learning rate is 0.001, and is divided by 2 every 100K gradient updates.
The network is trained with the Adam optimizer for 500K gradient updates.

\begin{table}[h]
\setlength{\tabcolsep}{3pt}
\centering
\small
\subfloat{
\begin{tabular}{llll}
\multicolumn{4}{c}{\textbf{Image Patch Encoder}} \\
\toprule
Layer      & Resolution  & Stride  & Norm./Act.  \\
\midrule
Input       & 16x16x3       &          &              \\
Conv 3x3    & 16x16x32    & 1        & ReLU      \\
Maxpool 2x2 & 8x8x32      & 2         &           \\
Conv 3x3    & 8x8x64      & 1        & GN(4)/ReLU      \\
Maxpool 2x2 & 4x4x64      & 2         &           \\
Conv 3x3    & 4x4x128      & 1        & GN(8)/ReLU      \\
Maxpool 2x2 & 2x2x128      & 2         &           \\
Conv 3x3    & 2x2x256      & 1        & GN(16)/ReLU      \\
Maxpool 2x2 & 1x1x256      & 2         &           \\
\bottomrule
\end{tabular}
}
\subfloat{
\adjustbox{raise=2.25em}{
\begin{tabular}{lll}
\multicolumn{3}{c}{\textbf{GNN}}   \\
\toprule
Layer   & Resolution    & Norm./Act. \\
\midrule
Input   & $N$x256     &  \\
Global Maxpool  & 256   &   \\
Linear (no bias)    & 512     & ReLU  \\
Linear (no bias)    & 512     & ReLU  \\
Linear      & 1     \\
\bottomrule
\end{tabular}
}
}
\vspace{0.5em}
\caption{
The architecture of GNN and image patch encoder used in the experiments on Multi-MNIST. $N$ denotes the number of nodes in a graph. Note that we use an empty graph here.}
\label{tab:arch-gnn-multi-mnist}
\end{table}

\subsection{Baselines}
\paragraph{Network Architectures}
Table~\ref{tab:arch-cnn-multi-mnist} shows the architecture of the plain CNN used in the experiments on Multi-MNIST.
The Relation Net shares the same CNN backbone with the plain CNN.
For the Relation Net, following \cite{zambaldi2018deep}, we add a relation module after the final feature map by a residual connection.
The architecture of the relation module is illustrated in Table~\ref{tab:arch-relation-multi-mnist}.
For these baselines, the last CNN feature map is flattened and a multi-layer perceptron (MLP) is applied to get the final output.
Other design choices, \eg, max pooling over the last feature map, are investigated and the quantitative results are shown in Table~\ref{tab:design-choices-multi-mnist}.
\paragraph{Hyperparameters in Tranining}
All the baseline CNNs are trained with the Adam optimizer for 500K steps.
The initial learning rate is also 0.001, and is divided by 2 every 100K steps.

\begin{table}[H]
\centering
\small
\begin{tabular}{llll}
\toprule
Layer       & Resolution   & Stride   & Norm./Act.   \\
\midrule
Input       & 54x54x3      &          &              \\
Conv 3x3    & 54x54x64     & 1        & BN/ReLU      \\
Conv 3x3    & 18x18x64      & 3        & BN/ReLU      \\
Conv 3x3    & 18x18x128    & 1        & BN/ReLU      \\
Conv 3x3    & 6x6x128      & 3        & BN/ReLU      \\
Conv 3x3    & 6x6x256      & 1        & BN/ReLU      \\
\midrule
Flatten     & 9216  \\
Linear (no bias)    & 512   &   & ReLU \\
Linear (no bias)    & 512   &   & ReLU \\
Linear      & 1 \\
\bottomrule
\end{tabular}
\vspace{0.5em}
\caption{The architecture of the plain CNN used in the experiments on Multi-MNIST.}
\label{tab:arch-cnn-multi-mnist}
\end{table}

\begin{table}[H]
\centering
\setlength{\tabcolsep}{3pt}
\small
\subfloat{
\begin{tabular}{llll}
\toprule
Layer       & Resolution   & Stride   & Norm./Act.   \\
\midrule
\multicolumn{4}{c}{\textbf{Key encoder}} \\
\midrule
Input       & 6x6x256      &          &              \\
Conv 1x1    & 6x6x256      & 1        & GN(4)/ReLU      \\
Conv 1x1    & 6x6x64      & 1        &       \\
\midrule
\multicolumn{4}{c}{\textbf{Query encoder}} \\
\midrule
Input       & 6x6x256      &          &              \\
Conv 1x1    & 6x6x256      & 1        & GN(4)/ReLU      \\
Conv 1x1    & 6x6x64      & 1        &       \\
\bottomrule
\end{tabular}
}
\hfill
\subfloat{
\begin{tabular}{llll}
\toprule
Layer       & Resolution   & Stride   & Norm./Act.   \\
\midrule
\multicolumn{4}{c}{\textbf{Value encoder}} \\
\midrule
Input       & 6x6x256      &          &              \\
Conv 1x1    & 6x6x256      & 1        & GN(4)/ReLU      \\
Conv 1x1    & 6x6x256      & 1        & GN(4)/ReLU      \\
\midrule
\multicolumn{4}{c}{\textbf{Post-attention Encoder}} \\
\midrule
Input       & 6x6x256      &          &              \\
Conv 1x1    & 6x6x256      & 1        & GN(4)/ReLU      \\
Conv 1x1    & 6x6x256      & 1        & GN(4)/ReLU      \\
\bottomrule
\end{tabular}
}
\vspace{0.5em}
\caption{The architecture of the relational module (4 heads) of relation netwrok used in the experiments on Multi-MNIST.}
\label{tab:arch-relation-multi-mnist}
\end{table}

\begin{table}[H]
\centering
\small
\setlength{\tabcolsep}{3pt}
\renewcommand{\arraystretch}{1.1}
\subfloat[Multi-MNIST-CIFAR]{
    \begin{tabular}{lll}
    \toprule
    Method                     & Train Acc & Test Acc \\
    \midrule
    CNN(flatten)    & 92.0(1.7)          & 30.7(4.9)         \\
    \midrule
    CNN(max pooling)    & 90.3(1.5)           & 20.2(3.6)         \\
    CNN(sum pooling)    & 91.0(0.9)           & 5.7(3.4)         \\
    \bottomrule
    \end{tabular}
}
\hfill
\subfloat[Multi-MNIST-ImageNet]{
    \begin{tabular}{lll}
    \toprule
    Method                     & Train Acc & Test Acc \\ 
    \midrule
    CNN(flatten)    &  90.5(2.9)           & 12.0(2.1)         \\
    \midrule
    CNN(max pooling)    & 83.1(0.7)           & 9.74(0.4)         \\
    CNN(sum pooling)    & 84.2(0.9)           & 10.2(1.8)         \\
    \bottomrule
    \end{tabular}
}
\vspace{0.5em}
\caption{Quantitative results of different design choices for the plain CNN on Multi-MNIST. The average with the standard deviation (in the parentheses) over 5 trials is reported.}
\label{tab:design-choices-multi-mnist}
\end{table}

\section{FallingDigit}
\label{sec:exp-falling-digit-supp}

\subsection{Environment Details}
The foreground (digit) images are randomly selected from the MNIST dataset. For each digit, we only use one fixed image instance. The background images are either black or random selected from a subset in CIFAR-10 \cite{krizhevsky2009learning} dataset, and this subset contains 100 random selected images. All the foreground and background images are shared across the training and test environments.

\subsection{Demonstration Acquisition}
We train a CNN-based DQN to acquire the teacher policy, which is used to interact with the FallingDigit environment with three target digits to collect the demonstration dataset.
During the interaction, we use the greedy policy derived from the $Q$ function, i.e., $\pi(s)=\argmax_{a} Q(s,a)$.
The demonstration dataset includes 60,000 images and each image is labeled with $Q(s, a)$ for all actions calculated by the teacher policy.

\subsection{Self-supervised Object Detector}
The self-supervised object detector is trained on the collected demonstration dataset.
For the glimpse of each object, we apply the STN to crop a patch from the image according to the bounding box and resize it to 16x16.
Table~\ref{tab:arch-space-falling-digit} shows the architecture of the object detector used in the experiments.
For FallingDigit-Black, the background encoder and decoder are removed.
Table~\ref{tab:hyper-space-falling-digit} shows the hyperparameters of the object detector.

\begin{table}[H]
\centering
\setlength{\tabcolsep}{3pt}
\small
\begin{tabular}{llll}
\multicolumn{4}{c}{\textbf{Foreground Image Encoder}} \\
\toprule
Layer       & Resolution   & Stride   & Norm./Act.   \\
\midrule
Input       & 128x128x3     &          &              \\
Conv 3x3    & 128x128x16    & 1        & BN/ReLU      \\
Maxpool     & 64x64x16      & 2        &              \\
Conv 3x3    & 64x64x32      & 1        & BN/ReLU      \\
Maxpool     & 32x32x32      & 2        &              \\
Conv 3x3    & 32x32x64      & 1        & BN/ReLU      \\
Maxpool     & 16x16x64      & 2        &              \\
Conv 3x3    & 16x16x128     & 1        & BN/ReLU      \\
Conv 1x1    & 16x16x128     & 1        & BN/ReLU      \\
Conv 1x1    & 16x16x128     & 1        & BN/ReLU      \\
\midrule
\multirow{3}{*}{Conv 1x1}   & 16x16x1 (object presence $z^{pres}$)     & 1     & Sigmoid   \\
& 16x16x4 (bounding box mean $z^{where}$)     & 1     &    \\
& 16x16x4 (bounding box stdev $z^{where}$)     & 1     & Softplus   \\
\bottomrule
\end{tabular}

\subfloat{
\begin{tabular}{llll}
\multicolumn{4}{c}{\textbf{Glimpse Encoder}}   \\
\toprule
Layer   & Resolution    & Stride    & Norm./Act. \\
\midrule
Input       & 16x16x3         &           &            \\
Conv 1x1    & 16x16x16        & 1         & ReLU    \\
Maxpool     & 8x8x32        & 2         &   \\
Conv 1x1    & 8x8x32        & 1         & ReLU    \\
Maxpool     & 4x4x32        & 2         &   \\
Conv 1x1    & 4x4x64        & 1         & ReLU    \\
Maxpool     & 2x2x64        & 2         &   \\
Conv 1x1    & 2x2x128        & 1         & ReLU    \\
Maxpool     & 1x1x128        & 2         &   \\
\multirow{2}{*}{Linear}     & 50 (mean $z^{what}$)    \\
            & 50 (stdev $z^{what}$)    &  & Softplus  \\
\bottomrule
\end{tabular}
}
\hfill
\subfloat{
\begin{tabular}{llll}
\multicolumn{4}{c}{\textbf{Glimpse Decoder}}   \\
\toprule
Layer   & Resolution    & Stride    & Norm./Act. \\
\midrule
Input       & 1x1x50         &           &            \\
Deconv 2x2    & 2x2x128        & 2         & ReLU    \\
Conv 1x1    & 2x2x64        & 1         & ReLU    \\
Deconv 2x2    & 4x4x64        & 2         & ReLU    \\
Conv 1x1    & 4x4x32        & 1         & ReLU    \\
Deconv 2x2    & 8x8x32        & 2         & ReLU    \\
Conv 1x1    & 8x8x16        & 1         & ReLU    \\
Upsample 2x2 & 16x16x16     & 2         &   \\
Conv 1x1    & 16x16x4        & 1         &    \\
\bottomrule
\end{tabular}
}
\\
\subfloat{
\begin{tabular}{llll}
\multicolumn{4}{c}{\textbf{Background Image Encoder}} \\
\toprule
Layer      & Resolution  & Stride  & Norm./Act.  \\
\midrule
Input       & 128x128x3       &          &              \\
Conv 3x3    & 128x128x32      & 1        & BN/ReLU      \\
Maxpool 2x2 & 64x64x32      & 2          &       \\
Conv 3x3    & 64x64x32      & 1        & BN/ReLU      \\
Maxpool 2x2 & 32x32x32      & 2          &       \\
Conv 3x3    & 32x32x32      & 1        & BN/ReLU      \\
Maxpool 2x2 & 16x16x32      & 2          &       \\
Conv 3x3    & 16x16x32      & 1        & BN/ReLU      \\
Maxpool 2x2 & 8x8x32      & 2          &       \\
\bottomrule
\end{tabular}
}
\hfill
\subfloat{
\begin{tabular}{llll}
\multicolumn{4}{c}{\textbf{Background Image Decoder}}   \\
\toprule
Layer   & Resolution    & Stride    & Norm./Act. \\
\midrule
Input       & 8x8x32         &           &            \\
Deconv 2x2    & 16x16x32        & 2         & BN/ReLU    \\
Conv 3x3    & 16x16x32        & 1         & BN/ReLU    \\
Deconv 2x2    & 32x32x32        & 2         &BN/ReLU    \\
Conv 3x3    & 32x32x32        & 1         & BN/ReLU    \\
Deconv 2x2    & 64x64x32        & 2         &BN/ReLU    \\
Conv 3x3    & 64x64x32        & 1         & BN/ReLU    \\
Upsample 2x2 & 128x128x32     & 2         &   \\
Conv 3x3    & 128x128x32        & 1         & BN/ReLU   \\
Conv 1x1    & 128x128x3        & 1         &    \\
\bottomrule
\end{tabular}
}
\caption{
The architecture of the self-supervised object detector for all the experiments on FallingDigit.}
\label{tab:arch-space-falling-digit}
\end{table}

\begin{table}[H]
\centering
\small
\begin{tabular}{lll}
\toprule
Name                    & Value                           & Schedule    \\
\midrule
max iteration           & 100K                              &             \\
optimizer               & Adam                            &             \\
batch size              & 8        & \\
learning rate           & 1e-3                            &             \\
gradient clip           & 1.0                            &             \\
$z_{pres}$ prior        & $0.1 \to 0.005$       & $0 \to 50K$ \\
$z_{pres}$ temperature  & $2.5 \to 0.5$       &  $0 \to 50K$ \\
$z_{where}$ prior mean  & $0$                 &             \\
$z_{where}$ prior stdev & $0.2$            &             \\
$z_{what}$ prior mean   & $0$                 &             \\
$z_{what}$ prior stdev  & $1.0$            &             \\
$z_{what}$ dimension    & 50            &             \\
$z_{depth}$ prior mean   & $0$                 &             \\
$z_{depth}$ prior stdev  & $1.0$            &             \\
$z_{depth}$ scale       & $10.0$            &             \\
fg recon prior stdev    & $0.15$          &  \\
bg recon prior stdev    & $0.1$ (Black) / $0.15$ (CIFAR)           &  \\
\bottomrule
\end{tabular}
\vspace{0.5em}
\caption{The hyperparameters of the self-supervised object detector for all the experiments on FallingDigit. Note that the background module is disabled for FallingDigit-Black.}
\label{tab:hyper-space-falling-digit}
\end{table}

\subsection{Policy GNN}
\paragraph{Network Architectures}
In the experiments on FallingDigit, the policy GNN is implemented as EdgeConv~\cite{dgcnn} in PyTorch Geometric~\cite{Fey/Lenssen/2019}.
The input graph is a complete graph, i.e., the edge set is $\{(i,j)|i,j \in\{1..n\}\}$ including self-loops, where $i,j$ are node indices. Each node corresponds to a detected object and the node feature includes an embedded image feature $x_{img}$ and the bounding box of the object $x_{box}$. To get $x_{img}$, we crop an image patch from the original image according to the bounding box, and resize it to 16x16, and then encode it by an image patch encoder. We concatenate $x_{img}$ and $x_{box}$ to get the node features and pass them into the GNN. The edge feature is the concatenation of the feature of the sender node, and the difference between the features of the sender and receiver nodes.
Table~\ref{tab:arch-gnn-falling-digit} shows the architecture of GNN and image patch encoder used in FallingDigit experiments.

\paragraph{Hyperparameters in Refactorization}
When training the GNN, the batch size is 64.
The initial learning rate is 0.001, and is divided by 2 every 100K gradient updates.
The network is trained with the Adam optimizer for 200K gradient updates.

\begin{table}[h]
\centering
\setlength{\tabcolsep}{2pt}
\small
\subfloat{
\begin{tabular}{llll}
\multicolumn{4}{c}{\textbf{Image Patch Encoder}} \\
\toprule
Layer      & Resolution  & Stride  & Norm./Act.  \\
\midrule
Input       & 16x16x3       &          &              \\
Conv 3x3    & 16x16x16      & 1        & ReLU      \\
Maxpool 2x2 & 8x8x16        & 2     &   \\
Conv 3x3    & 8x8x32      & 1        & ReLU      \\
Maxpool 2x2 & 4x4x32        & 2     &   \\
Conv 3x3    & 4x4x64      & 1        & GN(4)/ReLU      \\
Maxpool 2x2 & 2x2x64        & 2     &   \\
Conv 3x3    & 2x2x128      & 1        & GN(8)/ReLU      \\
Maxpool 2x2 & 1x1x128        & 2     &   \\
\bottomrule
\end{tabular}
}
\hfill
\subfloat{
\begin{tabular}{lll}
\multicolumn{3}{c}{\textbf{GNN}}   \\
\toprule
Layer                       & Resolution & Norm./Act. \\
\midrule
Input                       & $N$x(128+4)         &  \\
Message Passing             & $E$x(128+4)x2         & \\
Linear                      & $E$x128        & GN(8)/ReLU    \\
Linear                      & $E$x128        & GN(8)/ReLU    \\
Max Aggregation             & $N$x128         &   \\
Linear                      & $N$x128        & GN(8)/ReLU    \\
Linear                      & $N$x128        & GN(8)/ReLU    \\
Global Maxpool              & 128       & \\
Linear                      & 128        & ReLU    \\
Linear                      & 128        & ReLU    \\
Linear                      & 3        &     \\
\bottomrule
\end{tabular}
}
\vspace{0.5em}
\caption{
The architecture of GNN and image patch encoder used in FallingDigit. $N$ denotes the number of nodes in a graph, and $E$ denotes the number of edges in a graph. We use complete graph here.}
\label{tab:arch-gnn-falling-digit}
\end{table}

\subsection{Baselines}
\paragraph{Network Architectures}
\label{sec:baseline-arch-falling-digit}
The architecture of plain CNN is illustrated in Table~\ref{tab:arch-cnn-falling-digit}.
For the Relation Net \cite{zambaldi2018deep}, we follow most of the design choices described in the original paper.
We add a relational module at the 8x8 feature map. The relation module is the same as that used in Multi-MNIST, except that the resolution is 8x8.

\begin{table}[ht]
\centering
\setlength{\tabcolsep}{3pt}
\small
\begin{tabular}{llll}
\toprule
Layer       & Resolution   & Stride   & Norm./Act.   \\
\midrule
Input       & 128x128x3     &          &              \\
Conv 3x3    & 128x128x16    & 1        &  ReLU     \\
Maxpool     & 64x64x16      & 2        &              \\
Conv 3x3    & 64x64x16      & 1        &  ReLU     \\
Maxpool     & 32x32x16      & 2        &              \\
Conv 3x3    & 32x32x32      & 1        &  ReLU   \\
Maxpool     & 16x16x32      & 2        &              \\
Conv 3x3    & 16x16x64     & 1        &  ReLU     \\
Maxpool     & 8x8x64     & 2        &              \\
Conv 3x3    & 8x8x128     & 1        &   ReLU    \\
Maxpool     & 4x4x128      & 2        &              \\
Conv 1x1    & 4x4x128     & 1        &  ReLU     \\
Global Maxpool  & 128 \\
Linear  & 256   &   & ReLU \\
Linear  & 3   &   & ReLU \\
\bottomrule
\end{tabular}
\caption{The architecture of plain CNN used in FallingDigit.}
\label{tab:arch-cnn-falling-digit}
\end{table}

\paragraph{Hyperparameters in Tranining}
We use DQN \cite{mnih2015human} to train all the baselines. The related hyperparameters are listed in the Table \ref{tab:hyper-dqn-falling-digit}.

\begin{table}[ht]
\centering
\begin{tabular}{lll}
\toprule
Name                    & Value                           & Schedule    \\
\midrule
max iteration           & 10M                              &             \\
optimizer               & Adam                            &             \\
learning rate           & 1e-4                            &             \\
gradient clip           & 10.0                            &             \\
$\epsilon$-greedy       & $1.0 \to 0.1$       & $0 \to 300K$ \\
image normalizer        & divide by 255                &             \\
stacked frames        & 1                &             \\
target net update frequency & 500 steps                &             \\
replay buffer size & 100K                &             \\
discount factor & 0.99                &             \\
training frequency & 4 steps                &             \\
batch size & 32                &             \\
double Q & Yes                &             \\
\bottomrule
\end{tabular}
\vspace{0.5em}
\caption{The hyperparameters for training DQN on FallingDigit.}
\label{tab:hyper-dqn-falling-digit}
\end{table}

\subsection{Evaluation Method}
We train our GNN-based policy and all the baselines on the environment with three target digits and test them on the environments with more target digits. When evaluating all the policies, we take the best action suggested by the policy, i.e., $\pi_{eval}(s)=\argmax_a Q(s,a)$. Since the environments are stochastic (the positions of all digits are random), we evaluate every policy on every environment for 1000 episodes and calculate the mean episode reward.

\section{BigFish}
\label{sec:exp-bigfish-supp}
\subsection{Environment Details}
\begin{figure}[h]
\centering
\subfloat{\includegraphics[width=0.2\textwidth]{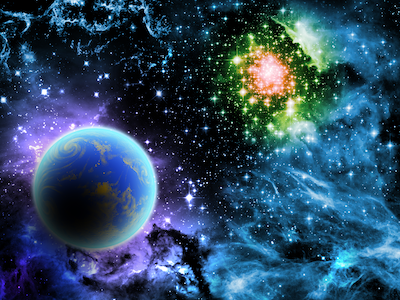}}\hfill
\subfloat{\includegraphics[width=0.2\textwidth]{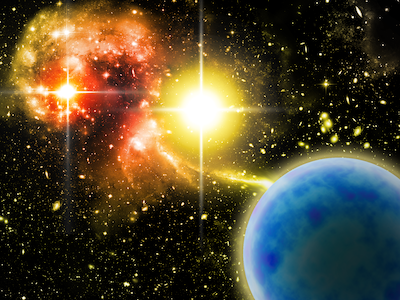}}\hfill
\subfloat{\includegraphics[width=0.2\textwidth]{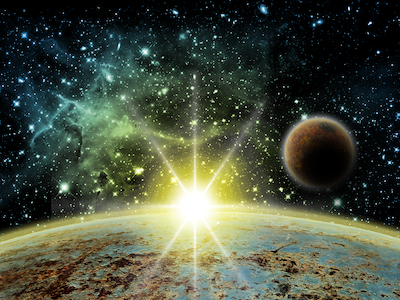}}\hfill
\subfloat{\includegraphics[width=0.2\textwidth]{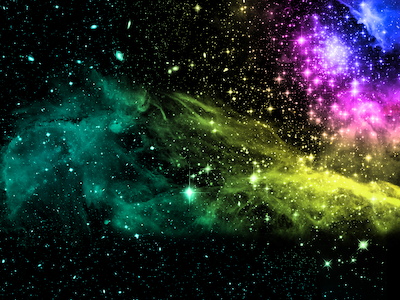}}\hfill
\caption{Background images used in BigFish test environments.}
\label{fig:bigfish-new-bg}
\end{figure}

Different from the original BigFish game, we modify the source code to add new background images to create test environments. To be more specific, we use the "Background-1.png", "Background-2.png", "Background-3.png" and "Background-4.png" from the "space-backgounds" directory in the ProcGen~\cite{cobbe2019procgen} source code. These background images are shown in Fig~\ref{fig:bigfish-new-bg}.

\subsection{Demonstration Acquisition}
To make our RL agent perform well in the test environment, it is important to train it on a diverse set of states which can cover most of states that might be encountered in the environments. In the BigFish environment, we found that if we simply use the best action suggested by the teacher policy, the demonstration dataset would only cover the states along the optimal trajectories, which are only a small portion of all feasible states. For example, in the optimal trajectories, the player will never get too close to the fishes that are bigger than the player itself. But in the testing environment, this may happen and our agent needs to know how to react in these states.

Therefore, we introduce $\epsilon$-greedy exploration \cite{mnih2015human} to increase the diversity of states. Since our teacher policy is trained by PPO \cite{schulman2017proximal}, the output of the teacher policy is a categorical distribution over 15 discrete actions. When we apply $\epsilon$-greedy strategy, the agent will take a random action with probability $\epsilon$, otherwise, the agent will take a action suggested by the teacher policy (sampled from the categorical distribution). We collect demonstration from every level in the training set (level 0 to 200), and for each level we run the teacher policy multiple times.

We combine the datasets collected with different values of $\epsilon$ and the details are listed in Table~\ref{tab:bigfish_dataset}. Since we run the experiments several times and collect several demonstration datasets, the size of the datasets vary. But we report a rough number of the sizes. The size of our combined demonstration dataset is around 800K.

\begin{table}[ht]
\centering
\begin{tabular}{lll}
\toprule
$\epsilon$    & \# of trials in each level     & dataset size \\
\midrule
0.5         & 5   &  $\sim$ 350k     \\
0.3         & 3   &  $\sim$ 300k        \\
0           & 1   &  $\sim$ 150k        \\
\bottomrule
\end{tabular}
\vspace{2mm}
\caption{Composition of the demonstration dataset for BigFish}
\label{tab:bigfish_dataset}
\end{table}

\subsection{Self-supervised Object Detector}
The self-supervised object detector is trained on a subset of the collected demonstration dataset, which consists of 60,000 images.
The setting is similar to that of Multi-MNIST.
For the glimpse of each object, we apply the STN to crop a patch from the image according to the bounding box and resize it to 8x8.
Table~\ref{tab:arch-space-bigfish} shows the architecture of the object detector used in the experiments.
Table~\ref{tab:hyper-space-bigfish} shows the hyperparameters of the object detector.

\begin{table}[ht]
\small
\centering
\begin{tabular}{llll}
\multicolumn{4}{c}{\textbf{Foreground Image Encoder}} \\
\toprule
Layer       & Resolution   & Stride   & Norm./Act.   \\
\midrule
Input       & 64x64x3       &          &              \\
Conv 3x3    & 64x64x64      & 1        & BN/ReLU      \\
Maxpool 2x2    & 32x32x64      & 2        &      \\
Conv 3x3    & 32x32x128      & 1        & BN/ReLU      \\
Maxpool 2x2    & 16x16x128      & 2        &      \\
Conv 3x3    & 16x16x256     & 1        & BN/ReLU      \\
Maxpool 2x2    &8x8x256      & 2        &      \\
Conv 1x1    & 8x8x256     & 1        & BN/ReLU      \\
Conv 1x1    & 8x8x256     & 1        & BN/ReLU      \\
\midrule
\multirow{3}{*}{Conv 1x1}   & 8x8x1 (object presence $z^{pres}$)     & 1     & Sigmoid   \\
& 8x8x4 (bounding box mean $z^{where}$)     & 1     &    \\
& 8x8x4 (bounding box stdev $z^{where}$)     & 1     & Softplus   \\
\bottomrule
\end{tabular}

\subfloat{
\begin{tabular}{lll}
\multicolumn{3}{c}{\textbf{Glimpse Encoder}}   \\
\toprule
Layer                       & Resolution & Norm./Act. \\
\midrule
Input                       & 8x8x3    &            \\
Flatten                     & 192        &            \\
Linear                      & 256        & GN(16)/ReLU    \\
Linear                      & 256        & GN(16)/ReLU    \\
\multirow{2}{*}{Linear}     & 32        &   \\
                            & 32    & Softplus  \\
\bottomrule
\end{tabular}
}
\hfill
\subfloat{
\begin{tabular}{lll}
\multicolumn{3}{c}{\textbf{Glimpse Decoder}}   \\
\toprule
Layer                       & Resolution & Norm./Act. \\
\midrule
Input                       & 32         &  \\
Linear                      & 256        & GN(16)/ReLU    \\
Linear                      & 256        & GN(16)/ReLU    \\
Linear                      & 256        & Sigmoid      \\
Reshape                     & 8x8x4    &   \\
\bottomrule
\end{tabular}
}
\\
\subfloat{
\begin{tabular}{llll}
\multicolumn{4}{c}{\textbf{Background Image Encoder}} \\
\toprule
Layer      & Resolution  & Stride  & Norm./Act.  \\
\midrule
Input       & 64x64x3       &          &              \\
Conv 3x3    & 64x64x64      & 1        & BN/ReLU      \\
Maxpool 2x2    & 32x32x64      & 2        &      \\
Conv 3x3    & 32x32x128      & 1        & BN/ReLU      \\
Maxpool 2x2    & 16x16x128      & 2        &      \\
Conv 3x3    & 16x16x256     & 1        & BN/ReLU      \\
Maxpool 2x2    &8x8x256      & 2        &      \\
Conv 1x1    & 8x8x256     & 1        & BN/ReLU      \\
\bottomrule
\end{tabular}
}
\hfill
\subfloat{
\begin{tabular}{lll}
\multicolumn{3}{c}{\textbf{Background Image Decoder}}  \\
\toprule
Layer      & Resolution  & Norm./Act.  \\
\midrule
Input       & 256       &   \\
Linear     & 256         & BN/ReLU     \\
Linear     & 256         & BN/ReLU     \\
Linear     & 12288        & Sigmoid      \\
Reshape    & 64x64x3     &     \\
\bottomrule
\end{tabular}
}
\vspace{0.5em}
\caption{
The architecture of the self-supervised object detector for all the experiments on BigFish.}
\label{tab:arch-space-bigfish}
\end{table}

\begin{table}[H]
\centering
\small
\begin{tabular}{lll}
\toprule
Name                    & Value                           & Schedule    \\
\midrule
max iteration           & 100K                              &             \\
optimizer               & Adam                            &             \\
batch size              & 32        & \\
learning rate           & 1e-3                            &             \\
gradient clip           & 1.0                            &             \\
$z_{pres}$ prior        & $0.15 \to 0.05$       & $10K \to 50K$ \\
$z_{pres}$ temperature  & $2.5 \to 0.5$       &  $10K \to 50K$ \\
$z_{where}$ prior mean  & $0$                 &             \\
$z_{where}$ prior stdev & $0.3$            &             \\
$z_{what}$ prior mean   & $0$                 &             \\
$z_{what}$ prior stdev  & $1.0$            &             \\
$z_{what}$ dimension    & 32            &             \\
$z_{depth}$ prior mean   & $0$                 &             \\
$z_{depth}$ prior stdev  & $1.0$            &             \\
$z_{depth}$ scale       & $10.0$            &             \\
fg recon prior stdev    & $0.15$          &  \\
bg recon prior stdev    & $0.15$          &  \\
\bottomrule
\end{tabular}
\vspace{0.5em}
\caption{The hyperparameters of the self-supervised object detector for all the experiments on BigFish.}
\label{tab:hyper-space-bigfish}
\end{table}

\subsection{Data Augmentation}
We find that data augmentation is helpful to the generalization of object-centric GNN policy in the BigFish environment. Specifically, the detection threshold in SPACE is usually set 0.1 in this paper. But here we randomly select 30\% of the object proposals (around 18 proposals) the has lower confidence score than the detection threshold and add them to the detection results. This data augmentation trick makes our object-centric GNN policy more robust to the false positive detections in the test environments. And this kind of data augmentation is not feasible for the CNN and Relation Net baselines.

\subsection{Policy GNN}
\paragraph{Network Architectures}
In the BigFish experiment, the policy GNN is implemented as EdgeConv~\cite{dgcnn} in PyTorch Geometric~\cite{Fey/Lenssen/2019}. The input graph is a complete graph, i.e., the edge set is $\{(i,j)|i,j \in\{1..n\}, i\neq j\}$, where $i,j$ are node indices. Each node corresponds to a detected object and the node feature includes an embedded image feature $x_{img}$ and the position of the object $x_{pos}$. To get $x_{img}$, we crop a 12x12 image patch from the original image according to the position of the object and encode it by an image patch encoder. And the $x_{pos}$ is a 2-dim vector which represents the coordinates of the center of the object bounding box. We concatenate $x_{img}$ and $x_{pos}$ to get the node features and pass them into the GNN. The edge feature is the concatenation of the feature of the sender node, and the difference between the features of the sender and receiver node.
Table~\ref{tab:arch-gnn-bigfish} shows the detailed architecture of GNN and image patch encoder used in BigFish experiments.

\begin{table}[ht]
\small
\centering

\subfloat{
\begin{tabular}{llll}
\multicolumn{4}{c}{\textbf{Image Patch Encoder}} \\
\toprule
Layer      & Resolution  & Stride  & Act.  \\
\midrule
Input       & 12x12x3       &          &              \\
Conv 3x3    & 10x10x32      & 1        & ReLU      \\
Conv 3x3    & 8x8x64      & 1        & ReLU      \\
Conv 8x1    & 1x8x64      & 1        & ReLU      \\
Flatten & 512      &            &       \\
Linear    & 512      &         & ReLU      \\
Linear    & 512      &         &       \\
\bottomrule
\end{tabular}
}
\hfill
\subfloat{
\begin{tabular}{lll}
\multicolumn{3}{c}{\textbf{GNN}}   \\
\toprule
Layer                       & Resolution & Act. \\
\midrule
Input                       & $N$x(512+2)         &  \\
Message Passing                      & $E$x(514x2)        & ReLU    \\
Linear                      & $E$x1024        & ReLU    \\
Linear                      & $E$x512        &       \\
Max Aggregation             & $N$x512       & \\
Global Maxpool           & 512       & \\
Linear                      & 15        &       \\
\bottomrule
\end{tabular}
}
\vspace{0.5em}
\caption{
The architecture of GNN and image patch encoder used in BigFish. $N$ denotes the number of nodes in a graph, and $E$ denotes the number of edges in a graph. We use complete graph here.}
\label{tab:arch-gnn-bigfish}
\end{table}

\paragraph{Hyperparameters in Refactorization}
When training the GNN, the batch size is 128.
The initial learning rate is 8e-4, reduced to 8e-5 at the 560K-th gradient updates, and then reduced to 8e-6 at the 750K-th gradient updates. The network is trained with the Adam optimizer for 1150K gradient updates.

\subsection{Baselines}
\paragraph{Network Architectures}
The CNN baseline is implemented according to the CNN architecture used in IMPALA \cite{impala}, which is suggested by the ProcGen paper \cite{cobbe2019procgen}.
For the Relation Net \cite{zambaldi2018deep} baseline, we use the same convolutional layers with IMPALA CNN, except we concatenate the spatial coordinates to the feature map as described in \cite{zambaldi2018deep}. Then, we add a relational module after the final feature map by a residual connection. The architecture of the relation module is illustrated in Table~\ref{tab:arch-relation-bigfish}. The output module of Relation Net is a flatten operator followed by a 2-layer MLP with hidden units of 256.

\begin{table}[ht]
\centering
\setlength{\tabcolsep}{3pt}
\small
\subfloat{
\begin{tabular}{llll}
\toprule
Layer       & Resolution   & Stride   & Norm./Act.   \\
\midrule
\multicolumn{4}{c}{\textbf{Key encoder}} \\
\midrule
Input       & 8x8x64      &          &              \\
Conv 1x1    & 8x8x64      & 1        & LN/ReLU      \\
Conv 1x1    & 8x8x64      & 1        &       \\
\midrule
\multicolumn{4}{c}{\textbf{Query encoder}} \\
\midrule
Input       & 8x8x64      &          &              \\
Conv 1x1    & 8x8x64      & 1        & LN/ReLU      \\
Conv 1x1    & 8x8x64      & 1        &       \\
\bottomrule
\end{tabular}
}
\hfill
\subfloat{
\begin{tabular}{llll}
\toprule
Layer       & Resolution   & Stride   & Norm./Act.   \\
\midrule
\multicolumn{4}{c}{\textbf{Value encoder}} \\
\midrule
Input       & 8x8x64      &          &              \\
Conv 1x1    & 8x8x64      & 1        & LN/ReLU      \\
Conv 1x1    & 8x8x64      & 1        & LN/ReLU      \\
\midrule
\multicolumn{4}{c}{\textbf{Post-attention Encoder}} \\
\midrule
Input       & 8x8x64      &          &              \\
Conv 1x1    & 8x8x64      & 1        & LN/ReLU      \\
Conv 1x1    & 8x8x64      & 1        & LN/ReLU      \\
\bottomrule
\end{tabular}
}
\vspace{0.5em}
\caption{The architecture of the relational module of Relation Net used in the experiments on BigFish. LN indicates Layer Normalization \cite{ba2016layer}.}
\label{tab:arch-relation-bigfish}
\end{table}

\paragraph{Hyperparameters in Training}
We use PPO \cite{schulman2017proximal} to train all the baselines and use the same hyperparameters with the ProcGen paper \cite{cobbe2019procgen}, except that we use the easy mode of the game and train 200M frames.

\subsection{Evaluation Method}
We train our GNN-based policy and all the baselines on level 0-199 and test them on level 200- 399. When evaluating all the policies, we take the best action suggested by the policy, i.e., $\pi_{eval}(s)=\argmax_a \pi(a|s)$ , instead of sampling from the categorical distribution. We evaluate every policy on every level from 200 to 399 once and calculate the mean episode reward. Since the environment is deterministic given the level index, and the policies are also deterministic by taking argmax, evaluating once is sufficient.

\section{Pacman}
\label{sec:exp-pacman-supp}
\subsection{Demonstration Acquisition}
We train a CNN-based DQN to acquire the teacher policy, which is used to interact with the Pacman environment with two dots (food) to collect the demonstration dataset.
During the interaction, we use the greedy policy derived from the $Q$ function, i.e., $\pi(s)=\argmax_{a} Q(s,a)$.
The demonstration dataset includes 60,000 images and each image is labeled with $Q(s, a)$ for all actions calculated by the teacher policy.
According to our experiment results, this demonstration dataset is good enough for learning a GNN-based student policy which can generalize to the environments with more dots.

\subsection{Self-supervised Object Detector}
The self-supervised object detector is trained on the collected demonstration dataset.
The setting is similar to that of Multi-MNIST.
For the glimpse of each object, we apply the STN to crop a patch from the image according to the bounding box and resize it to 8x8.
Table~\ref{tab:arch-space-pacman} shows the architecture of the object detector used in the experiments.
Table~\ref{tab:hyper-space-pacman} shows the hyperparameters of the object detector.

\begin{table}[ht]
\centering
\setlength{\tabcolsep}{3pt}
\small
\begin{tabular}{llll}
\multicolumn{4}{c}{\textbf{Foreground Image Encoder}} \\
\toprule
Layer       & Resolution   & Stride   & Norm./Act.   \\
\midrule
Input       & 64x64x3       &          &              \\
Conv 3x3    & 64x64x32      & 1        & BN/ReLU      \\
Conv 2x2    & 32x32x32      & 2        & BN/ReLU      \\
Conv 3x3    & 32x32x64      & 1        & BN/ReLU      \\
Conv 2x2    & 16x16x128     & 2        & BN/ReLU      \\
Conv 1x1    & 16x16x128     & 1        & BN/ReLU      \\
Conv 1x1    & 16x16x128     & 1        & BN/ReLU      \\
\midrule
\multirow{3}{*}{Conv 1x1}   & 16x16x1 (object presence $z^{pres}$)     & 1     & Sigmoid   \\
& 16x16x4 (bounding box mean $z^{where}$)     & 1     &    \\
& 16x16x4 (bounding box stdev $z^{where}$)     & 1     & Softplus   \\
\bottomrule
\end{tabular}

\subfloat{
\begin{tabular}{llll}
\multicolumn{4}{c}{\textbf{Glimpse Encoder}}   \\
\toprule
Layer   & Resolution    & Stride    & Norm./Act. \\
\midrule
Input       & 8x8x3         &           &            \\
Conv 1x1    & 8x8x32        & 1         & GN(4)/ReLU    \\
Maxpool     & 4x4x32        & 2         &   \\
Conv 1x1    & 4x4x64        & 1         & GN(4)/ReLU    \\
Maxpool     & 2x2x64        & 2         &   \\
Conv 1x1    & 2x2x128        & 1         & GN(8)/ReLU    \\
Maxpool     & 1x1x128        & 2         &   \\
\multirow{2}{*}{Linear}     & 32    \\
            & 32    &  & Softplus  \\
\bottomrule
\end{tabular}
}
\hfill
\subfloat{
\begin{tabular}{llll}
\multicolumn{4}{c}{\textbf{Glimpse Decoder}}   \\
\toprule
Layer   & Resolution    & Stride    & Norm./Act. \\
\midrule
Input       & 1x1x32         &           &            \\
Deconv 2x2    & 2x2x128        & 2         & GN(8)/ReLU    \\
Conv 1x1    & 2x2x64        & 1         & GN(4)/ReLU    \\
Deconv 2x2    & 4x4x64        & 2         & GN(4)/ReLU    \\
Conv 1x1    & 4x4x32        & 1         & GN(4)/ReLU    \\
Deconv 2x2    & 8x8x32        & 2         & GN(4)/ReLU    \\
Conv 1x1    & 8x8x16        & 1         & GN(4)/ReLU    \\
Conv 1x1    & 8x8x4        & 1         &    \\
\bottomrule
\end{tabular}
}
\\
\subfloat{
\begin{tabular}{llll}
\multicolumn{4}{c}{\textbf{Background Image Encoder}} \\
\toprule
Layer      & Resolution  & Stride  & Norm./Act.  \\
\midrule
Input       & 64x64x3       &          &              \\
Conv 3x3    & 64x64x32      & 1        & BN/ReLU      \\
Maxpool 2x2 & 32x32x32      & 2          &       \\
Conv 3x3    & 32x32x32      & 1        & BN/ReLU      \\
Maxpool 2x2 & 16x16x32      & 2          &       \\
Conv 3x3    & 16x16x32      & 1        & BN/ReLU      \\
Maxpool 2x2 & 8x8x32      & 2          &       \\
Conv 3x3    & 8x8x32      & 1        & BN/ReLU      \\
Maxpool 2x2 & 4x4x32      & 2          &       \\
\bottomrule
\end{tabular}
}
\hfill
\subfloat{
\begin{tabular}{llll}
\multicolumn{4}{c}{\textbf{Background Image Decoder}}   \\
\toprule
Layer   & Resolution    & Stride    & Norm./Act. \\
\midrule
Input       & 4x4x32         &           &            \\
Deconv 2x2    & 8x8x32        & 2         & BN/ReLU    \\
Conv 1x1    & 8x8x32        & 1         & BN/ReLU    \\
Deconv 2x2    & 16x16x32        & 2         & BN/ReLU    \\
Conv 1x1    & 16x16x32        & 1         & BN/ReLU    \\
Deconv 2x2    & 32x32x32        & 2         &BN/ReLU    \\
Conv 1x1    & 32x32x32        & 1         & BN/ReLU    \\
Deconv 2x2    & 64x64x32        & 2         &BN/ReLU    \\
Conv 1x1    & 64x64x32        & 1         & BN/ReLU    \\
Conv 1x1    & 64x64x3        & 1         &    \\
\bottomrule
\end{tabular}
}
\caption{
The architecture of the self-supervised object detector for all the experiments on Pacman.}
\label{tab:arch-space-pacman}
\end{table}

\begin{table}[ht]
\centering
\small
\begin{tabular}{lll}
\toprule
Name                    & Value                           & Schedule    \\
\midrule
max iteration           & 100K                              &             \\
optimizer               & Adam                            &             \\
batch size              & 8        & \\
learning rate           & 1e-3                            &             \\
gradient clip           & 1.0                            &             \\
$z_{pres}$ prior        & $0.1 \to 0.005$       & $0 \to 50K$ \\
$z_{pres}$ temperature  & $2.5 \to 0.5$       &  $0 \to 50K$ \\
$z_{where}$ prior mean  & $0$                 &             \\
$z_{where}$ prior stdev & $0.2$            &             \\
$z_{what}$ prior mean   & $0$                 &             \\
$z_{what}$ prior stdev  & $1.0$            &             \\
$z_{what}$ dimension    & 32            &             \\
$z_{depth}$ prior mean   & $0$                 &             \\
$z_{depth}$ prior stdev  & $1.0$            &             \\
$z_{depth}$ scale       & $10.0$            &             \\
fg recon prior stdev    & $0.15$          &  \\
bg recon prior stdev    & $0.15$          &  \\
\bottomrule
\end{tabular}
\vspace{0.5em}
\caption{The hyperparameters of the self-supervised object detector for all the experiments on Pacman.}
\label{tab:hyper-space-pacman}
\end{table}

\subsection{Policy GNN}
\paragraph{Network Architectures}
In the experiments on Pacman, the policy GNN is implemented as PointConv~\cite{qi2017pointnet++} in Pytorch Geometric~\cite{Fey/Lenssen/2019}.
The input graph is a complete graph, i.e., the edge set is $\{(i,j)|i,j \in\{1..n\}\}$ including self-loops, where $i,j$ are node indices. Each node corresponds to a detected object and the node feature includes an embedded image feature $x_{img}$ and the bounding box of the object $x_{box}$. To get $x_{img}$, we crop an image patch from the original image according to the bounding box, and resize it to 8x8, and then encode it by an image patch encoder. We concatenate $x_{img}$ and $x_{box}$ to get the node features and pass them into the GNN. The edge feature is the concatenation of the features of the sender node, and the difference between the bounding box position and size of the sender and receiver node.
Table~\ref{tab:arch-gnn-pacman} shows the architecture of GNN and image patch encoder used in Pacman experiments.

\paragraph{Hyperparameters in Refactorization}
When training the GNN, the batch size is 64.
The initial learning rate is 0.001, and is divided by 2 every 100K gradient updates.
The network is trained with the Adam optimizer for 500K gradient updates.

\begin{table}[H]
\centering
\setlength{\tabcolsep}{2pt}
\small
\subfloat{
\begin{tabular}{llll}
\multicolumn{4}{c}{\textbf{Image Patch Encoder}} \\
\toprule
Layer      & Resolution  & Stride  & Norm./Act.  \\
\midrule
Input       & 8x8x3       &          &              \\
Conv 3x3    & 8x8x32      & 1        & ReLU      \\
Maxpool 2x2 & 4x4x32        & 2     &   \\
Conv 3x3    & 4x4x64      & 1        & GN(4)/ReLU      \\
Maxpool 2x2 & 2x2x64        & 2     &   \\
Conv 3x3    & 2x2x128      & 1        & GN(8)/ReLU      \\
Maxpool 2x2 & 1x1x128        & 2     &   \\
\bottomrule
\end{tabular}
}
\hfill
\subfloat{
\begin{tabular}{lll}
\multicolumn{3}{c}{\textbf{GNN}}   \\
\toprule
Layer                       & Resolution & Norm./Act. \\
\midrule
Input                       & $N$x(128+4)         &  \\
Message Passing             & $E$x(128+4)         & \\
Linear                      & $E$x128        & GN(8)/ReLU    \\
Linear                      & $E$x128        & GN(8)/ReLU    \\
Linear                      & $E$x4        &       \\
Sum Aggregation             & $N$x4         &   \\
Global Maxpool              & 4       & \\
\bottomrule
\end{tabular}
}
\vspace{0.5em}
\caption{
The architecture of GNN and image patch encoder used in Pacman. $N$ denotes the number of nodes in a graph, and $E$ denotes the number of edges in a graph. We use complete graph here.}
\label{tab:arch-gnn-pacman}
\end{table}

\subsection{Baselines}
\paragraph{Network Architectures}
\label{sec:baseline-arch-pacman}
The architecture of plain CNN is illustrated in Table~\ref{tab:arch-cnn-pacman}.
For the Relation Net~\cite{zambaldi2018deep}, we follow most of the design choices described in the original paper. In our implementation, the input module of the Relation Net is the same as the convolutional layers used in the CNN baseline, except we concatenate the spatial coordinates to the feature map as described in \cite{zambaldi2018deep}. Then we add a relational module after the final feature map by a residual connection. The architecture of the relation module is illustrated in Table~\ref{tab:arch-relation-pacman}. The output module of Relation Net is a feature-wise max pooling layer followed by a 2-layer MLP with hidden units of 256.

\begin{table}[ht]
\centering
\setlength{\tabcolsep}{3pt}
\small
\begin{tabular}{llll}
\toprule
Layer       & Resolution   & Stride   & Norm./Act.   \\
\midrule
Input       & 64x64x3     &          &              \\
Conv 3x3    & 64x64x16      & 1        &  ReLU     \\
Maxpool     & 32x32x16      & 2        &              \\
Conv 3x3    & 32x32x32      & 1        &  ReLU   \\
Maxpool     & 16x16x32      & 2        &              \\
Conv 3x3    & 16x16x64     & 1        &  ReLU     \\
Maxpool     & 8x8x64     & 2        &              \\
Conv 3x3    & 8x8x128     & 1        &   ReLU    \\
Maxpool     & 4x4x128      & 2        &              \\
Conv 1x1    & 4x4x128     & 1        &  ReLU     \\
Global Maxpool  & 128 \\
Linear  & 256   &   & ReLU \\
Linear  & 4   &   & ReLU \\
\bottomrule
\end{tabular}
\vspace{0.5em}
\caption{The architecture of plain CNN used in Pacman.}
\label{tab:arch-cnn-pacman}
\end{table}

\begin{table}[ht]
\centering
\setlength{\tabcolsep}{3pt}
\small
\subfloat{
\begin{tabular}{llll}
\toprule
Layer       & Resolution   & Stride   & Norm./Act.   \\
\midrule
\multicolumn{4}{c}{\textbf{Key encoder}} \\
\midrule
Input       & 7x7x64      &          &              \\
Conv 1x1    & 7x7x64      & 1        & LN/ReLU      \\
Conv 1x1    & 7x7x64      & 1        &       \\
\midrule
\multicolumn{4}{c}{\textbf{Query encoder}} \\
\midrule
Input       & 7x7x64      &          &              \\
Conv 1x1    & 7x7x64      & 1        & LN/ReLU      \\
Conv 1x1    & 7x7x64      & 1        &       \\
\bottomrule
\end{tabular}
}
\hfill
\subfloat{
\begin{tabular}{llll}
\toprule
Layer       & Resolution   & Stride   & Norm./Act.   \\
\midrule
\multicolumn{4}{c}{\textbf{Value encoder}} \\
\midrule
Input       & 7x7x64      &          &              \\
Conv 1x1    & 7x7x64      & 1        & LN/ReLU      \\
Conv 1x1    & 7x7x64      & 1        & LN/ReLU      \\
\midrule
\multicolumn{4}{c}{\textbf{Post-attention Encoder}} \\
\midrule
Input       & 7x7x64      &          &              \\
Conv 1x1    & 7x7x64      & 1        & LN/ReLU      \\
Conv 1x1    & 7x7x64      & 1        & LN/ReLU      \\
\bottomrule
\end{tabular}
}
\vspace{0.5em}
\caption{The architecture of the relational module of Relation Net used in the experiments on Pacman. LN indicates Layer Normalization \cite{ba2016layer}.}
\label{tab:arch-relation-pacman}
\end{table}

\paragraph{Hyperparameters in Tranining}
We use DQN \cite{mnih2015human} to train all the baselines. The related hyperparameters are listed in the Table \ref{tab:hyper-dqn-pacman}.

\begin{table}[ht]
\centering
\begin{tabular}{lll}
\toprule
Name                    & Value                           & Schedule    \\
\midrule
max iteration           & 10M                              &             \\
optimizer               & Adam                            &             \\
learning rate           & 1e-4                            &             \\
gradient clip           & 10.0                            &             \\
$\epsilon$-greedy       & $1.0 \to 0.1$       & $0 \to 1M$ \\
image normalizer        & divide by 255                &             \\
stacked frames        & 1                &             \\
target net update frequency & 500 steps                &             \\
replay buffer size & 300K                &             \\
discount factor & 0.99                &             \\
training frequency & 4 steps                &             \\
batch size & 32                &             \\
double Q & Yes                &             \\
\bottomrule
\end{tabular}
\vspace{0.5em}
\caption{The hyperparameters for training DQN on Pacman.}
\label{tab:hyper-dqn-pacman}
\end{table}

\subsection{Evaluation Method}
We train our GNN-based policy and all the baselines on the environment with two dots and test them on the environments with more dots. When evaluating all the policies, we take the best action suggested by the policy, i.e., $\pi_{eval}(s)=\argmax_a Q(s,a)$. Since the environments are stochastic (the positions of Pacman and dots are random), we evaluate every policy on every environment for 100 episodes and calculate the mean episode reward.

\section{Robustness Analysis}
\label{sec:robustness-analysis}
In this section, we analyze the robustness of our two-stage refactorization framework, taking Pacman environment as an example. And we compare our two-stage refactorization framework with the end-to-end one-stage reinforcement learning method to show our framework is more robust to the low-quality detectors.

\subsection{Robustness w.r.t low-recall detectors}
\label{sec:robust-low-recall}
First, we conduct experiments to test how our refactorized GNN policy performs with a low-recall detector. Since the recall/AP of our object detector on Pacman is quite high, we randomly removed some detected objects to simulate the behaviours of a low-recall detector.

\begin{figure}[h]
\centering
\captionsetup[subfigure]{labelformat=empty}
\subfloat{\includegraphics[width=0.45\textwidth]{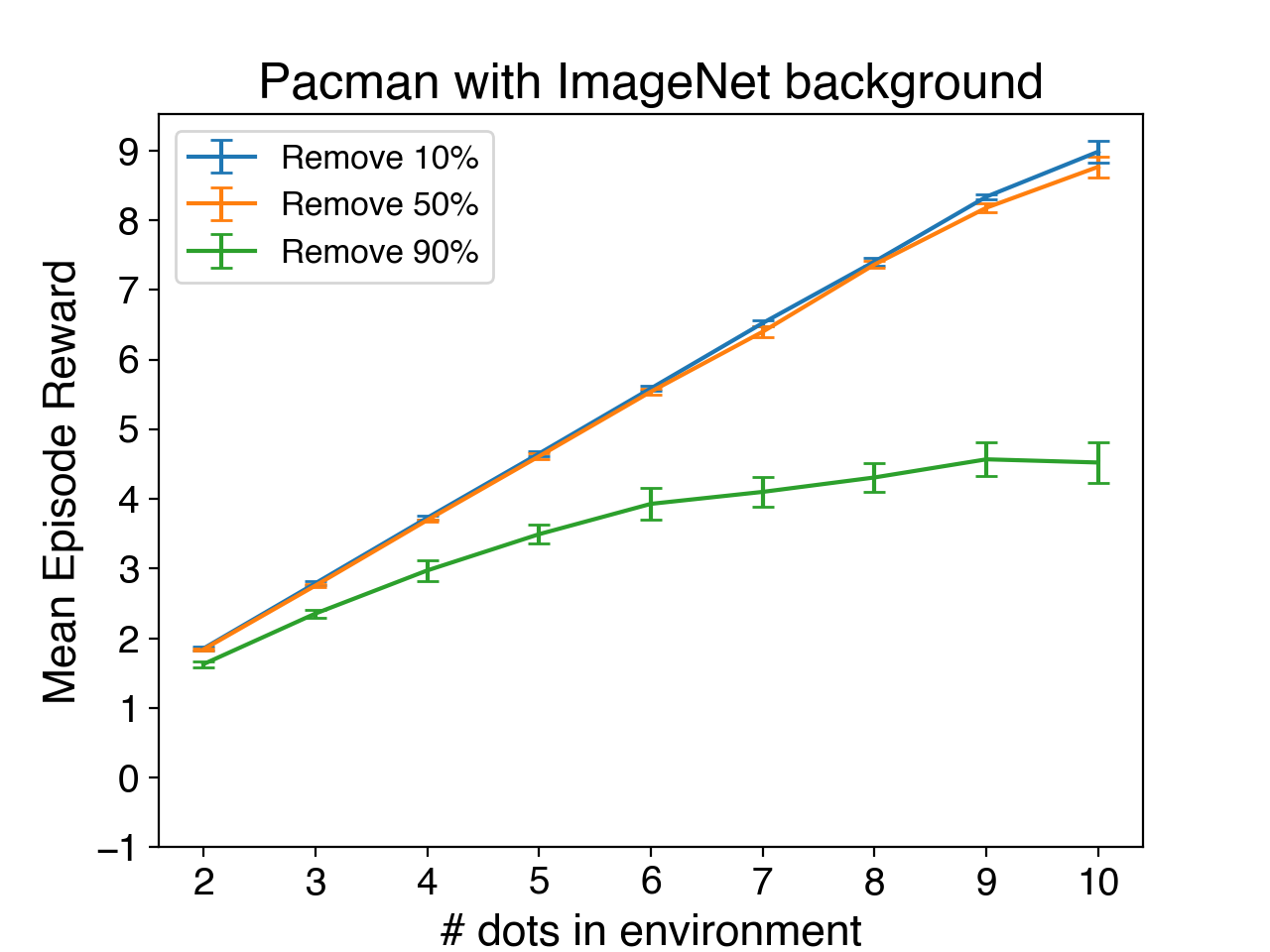}}
\caption{Quantitative experiments of robustness test on Pacman. We randomly remove 10\%, 50\% or 90\% detected objects during training and report the test performance in the environments with different number of dots.}
\label{fig:pacman-robustness}
\end{figure}

The detected objects are randomly removed in the demonstration dataset \textbf{during training} but are not removed \textbf{during testing}. We experiment with three different ratios of removed objects: 10\%, 50\% and 90\%.
Surprisingly, it is observed that even 50\% objects are removed, the policy GNN can also imitate a reasonable policy from the demonstration dataset, and still generalizes well to the environments with more dots.
We argue that it results from both the nature of the game itself and the robustness of our framework.
Fig~\ref{fig:pacman-robustness} illustrates the quantitative results.

In contrast, if we train our policy GNN with reinforcement learning (DQN) with 50\% objects missing, it cannot converges to a reasonable good solution.

\subsection{Robustness w.r.t low-precision detectors}
\label{sec:robust-low-precision}
Second, we test whether our refactorized GNN policy is robust to a low-precision detector. Similar to Sec~\ref{sec:robust-low-recall}, we simulate the behaviours of a low-precision detector by modifying a good detector. Specifically, we randomly select 25 object proposals with confidence scores lower than the threshold, which means they are not real objects, and add them to the detection results.

With such a low-precision detector, our refactorized GNN policy can still generalize well the environments with 10 dots (gets 8.29, averaged by 8 runs). In contrast, if we train our policy GNN with reinforcement learning (DQN) with this low-recall detector, the resulting GNN policy cannot consistently generalize to the environments with 10 dots (gets 4.43, averaged by 9 runs).

\subsection{Summary}
Through the above presented experiments, we find that our refactorized GNN is pretty robust w.r.t the low-recall detectors and low-precision detectors in the Pacman environment. In contrast, training the same object-centric GNN using reinforcement learning with low-quality detectors may lead to optimization or generalization problems. This is one of the reasons that we choose to break the policy learning problem into two stages instead of relying on end-to-end RL.

\newpage
\bibliography{references}
\bibliographystyle{plain}